\documentclass[journal]{IEEEtran}

\IEEEoverridecommandlockouts                              

\usepackage{booktabs}

\usepackage{amsmath, amssymb}
\usepackage{tikz}              
\usepackage{tikz-3dplot}
\usepackage{tikz-qtree}
\usepackage{cite}
\usepackage{algorithm}  
\usepackage{algpseudocode}
\usepackage{multirow}

\usepackage{amsmath}
\usepackage{amssymb}
\usepackage{amsmath} 

\usepackage[flushleft]{threeparttable}
\usepackage{subfigure}

\usepackage{tabularx} 
\DeclareGraphicsExtensions{.pdf,.jpg,.png}

\usepackage{graphicx}
\usepackage{caption}
\usepackage{float}
\usepackage[english]{babel}
\addto\captionsenglish{}

\usepackage{caption}
\usepackage{float}
\title{Scenario-based Decision-making Using Game Theory for Interactive Autonomous Driving: A Survey 
}

\author{Zhihao Lin$^{1}$, 
         Zhen Tian$^{1 \dagger}$, 
\thanks{$^{1}$Zhen Tian and Zhihao Lin are with the School of Engineering, University of Glasgow, Glasgow, G12 8QQ, U.K. (e-mail: 2620920z@student.gla.ac.uk, 2800400l@student.gla.ac.uk}%
\thanks{$\dagger$ Corresponding Author}
}

\begin{document}
\maketitle
\pagestyle{empty}  
\thispagestyle{empty} 


\begin{abstract}
Game-based interactive driving simulations have emerged as versatile platforms for advancing decision-making algorithms in road transport mobility. While these environments offer safe, scalable, and engaging settings for testing driving strategies, ensuring both realism and robust performance amid dynamic and diverse scenarios remains a significant challenge. Recently, the integration of game-based techniques with advanced learning frameworks has enabled the development of adaptive decision-making models that effectively manage the complexities inherent in varied driving conditions. These models outperform traditional simulation methods, especially when addressing scenario-specific challenges, ranging from obstacle avoidance on highways and precise maneuvering during on-ramp merging to navigation in roundabouts, unsignalized intersections, and even the high-speed demands of autonomous racing. Despite numerous innovations in game-based interactive driving, a systematic review comparing these approaches across different scenarios is still missing. This survey provides a comprehensive evaluation of game-based interactive driving methods by summarizing recent advancements and inherent roadway features in each scenario. Furthermore, the reviewed algorithms are critically assessed based on their adaptation of the standard game model and an analysis of their specific mechanisms to understand their impact on decision-making performance. Finally, the survey discusses the limitations of current approaches and outlines promising directions for future research.

\end{abstract}



\section{Introduction}\label{sec:intro}
The rapid growth in vehicle numbers worldwide has placed unprecedented pressure on existing transportation systems, making intelligent transportation solutions more crucial than ever~\cite{muratori2021rise,duarte2018impact}. With increasing urbanization and the consequent rise in traffic density, traditional infrastructure and management strategies struggle to keep pace with the demand for efficiency and safety~\cite{de2018itssafe,jagatheesaperumal2024artificial}. Intelligent transportation systems (ITS) are emerging as vital components of modern cities, integrating advanced communication, sensor networks, blocked chains, and data analytics to enhance traffic flow, reduce congestion, and improve overall mobility~\cite{10477849,10932698,lin2025conflicts,zuo2025industrial,fujiang2025ai,lin2025safety,zuo2025advanced,he2025integrated,yang2025application,ju2025research}. The Yolo-based networks have been widely used for the sensing stage~\cite{zuo2025intelligent,zhong2025skybound,qin2025ppa} As more vehicles crowd the roads, the complexity of interactions among road users escalates, demanding more adaptive and responsive solutions~\cite{10704959,xiong2023review}.

Traditional human-driven vehicles (HDVs) have long been the foundation of road transportation, yet they come with inherent drawbacks. Human drivers are susceptible to delayed reaction times, distractions, and cognitive biases that compromise their ability to respond swiftly and accurately to dynamic traffic situations~\cite{kyung2010enhancing,moser2010managing,salmon2019bad,BADUE2021113816}. These limitations are not just minor inconveniences; they are a major contributor to the high incidence of traffic accidents. Every year, a significant number of road injuries and fatalities can be traced back to human error~\cite{UKGov2023}. For example, in 2019, Germany recorded 300,143 road incidents that resulted in injuries, along with 3,046 fatalities~\cite{BASt2020}. Furthermore, over the past 30 years, more than 50,000 individuals have lost their lives on Australian roads~\cite{tao2022advanced}. Such alarming statistics underscore the critical need for alternative approaches that mitigate the risks associated with manual driving and improve road safety.

In response to these challenges, automated vehicles (AVs) have recently emerged as a transformative solution with the potential to revolutionize road transport. Equipped with sophisticated sensors~\cite{lin2024dpl,lin2025slam2,lin2024enhanced}, real-time data processing, and advanced decision-making algorithms, AVs offer a promising alternative by reducing reliance on human intervention. However, there are still challenges for AVs. Since 2021, more than 900 incidents involving Tesla's driver-assistance systems have been documented \cite{omeiza2021explanations}. Despite ongoing safety challenges, forecasts predict that the number of automated vehicles will exceed 50 million by 2024 \cite{ignatious2022overview}. These vehicles are designed to operate with greater consistency and precision, which is supposed to lead to enhanced driving quality, reduced congestion, and improved safety outcomes \cite{perez2011autonomous}. The promise of AVs lies in their ability to eliminate many of the pitfalls inherent in human driving, thereby creating a more efficient and secure transportation environment~\cite{ahmed2022technology,eskandarian2024advanced}.

Recently, the path planning of AVs has witnessed great development in non-interaction driving~\cite{yuan2025bio}. However, the integration of AVs into a predominantly human-driven traffic ecosystem introduces its own set of challenges~\cite{tan2021human,ezzati2021interaction}. Interactive decision-making becomes particularly complex when AVs must navigate environments dominated by unpredictable human behavior~\cite{wang2024towards,chen2024review,tian2025evaluating}. The erratic and varied nature of human driving styles requires decision-making systems in AVs to be exceptionally robust and adaptive. This interaction between AVs and HDVs creates a dynamic and often unpredictable landscape, where the success of AV algorithms hinges on their ability to respond intelligently to the actions of human drivers.

To address these challenges, researchers have explored a variety of interactive decision-making methods over the years. One widely studied approach is the potential field-based method, which draws inspiration from classical physics by representing goals as attractive forces and obstacles as repulsive forces~\cite{yao2020path,triharminto2017local}. This method is prized for its conceptual simplicity and its ability to compute smooth, continuous trajectories in real time. However, it is also known to suffer from issues like local minima, where the decision process may become trapped in a suboptimal state. For example, the classical Artificial Potential Field (APF) algorithm computes force vectors to steer the vehicle away from obstacles~\cite{triharminto2016novel,li2025adaptive}, while its enhanced variant, the Navigation Function method, incorporates additional heuristics to mitigate the local minima problem.

Another prominent strategy is robust control techniques, which focus on designing controllers that can tolerate uncertainties and variations in system dynamics~\cite{lofberg2003minimax}. These methods ensure that vehicle performance remains acceptable even when faced with unpredictable environmental disturbances or model inaccuracies. Yet, robust control techniques can sometimes lack the agility required to handle rapidly changing driving scenarios. For instance, H-infinity control has been effectively applied to maintain stability under varying conditions~\cite{glover2021h}, and Sliding Mode Control (SMC) has been employed to enforce precise trajectory tracking despite uncertainties~\cite{wu2021sliding}.

Sampling-based algorithms have gained attention due to their ability to explore a vast space of potential actions by generating numerous candidate trajectories and selecting the best one based on performance criteria~\cite{li2025efficient,tian2025risk}. These methods are particularly useful in complex, high-dimensional decision spaces where traditional optimization may be infeasible. However, the trade-off between computational efficiency and real-time performance remains a significant challenge. For example, Monte Carlo Tree Search (MCTS) has been adapted to evaluate future scenarios and select optimal maneuvers~\cite{browne2012survey,lenz2016tactical,lin2025multi,lin2025safetyr}, while Rapidly-exploring Random Trees (RRT) are often used to generate feasible paths in dynamic environments~\cite{xu2024recent,li2021adaptive}.

Rule-based systems represent a more traditional approach, relying on a set of predefined if-then rules to guide decision-making~\cite{bhattacharyya2021hybrid,xiao2021rule}. Their strength lies in their interpretability and ease of implementation, as engineers can directly incorporate expert knowledge and regulatory guidelines into the system. However, these systems often lack the flexibility to adapt to unforeseen scenarios and may become overly rigid. For instance, fuzzy logic-based rule systems combine multiple sensor inputs to generate decisions that mimic human reasoning, and finite state machines structure the decision process into distinct states with transitions triggered by specific environmental cues~\cite{bouchard2022rule}.

Finally, learning-based methods, particularly those harnessing deep learning and reinforcement learning, have emerged as powerful tools for developing adaptive decision-making strategies~\cite{9718218,li2018estimating,hu2025applications,zhen2025balanced}. These approaches can learn complex behaviors from large datasets or through trial and error in simulation environments, enabling vehicles to handle intricate and dynamic driving situations. Yet, learning-based methods sometimes encounter issues such as overfitting and a lack of interpretability, making it difficult to understand the underlying decision processes. As examples, Deep Q-Networks (DQN) have been employed to optimize vehicle maneuvers in simulation, and actor-critic architectures have been deployed to effectively balance exploration and exploitation in real-time decision-making~\cite{10214640}. Additionally, imitation learning techniques have been explored, where models learn to replicate expert driver behavior, further enhancing decision-making performance~\cite{cai2021vision}. Collectively, these approaches illustrate the diverse strategies researchers have deployed to enhance interactive decision-making in mixed traffic environments. Each method brings its unique strengths and limitations to the table, underscoring the need for continued innovation in order to achieve the agility, safety, and adaptability required for next-generation autonomous driving systems.
\begin{figure*}[ht]
    \centering
    \includegraphics[width=0.9\linewidth]{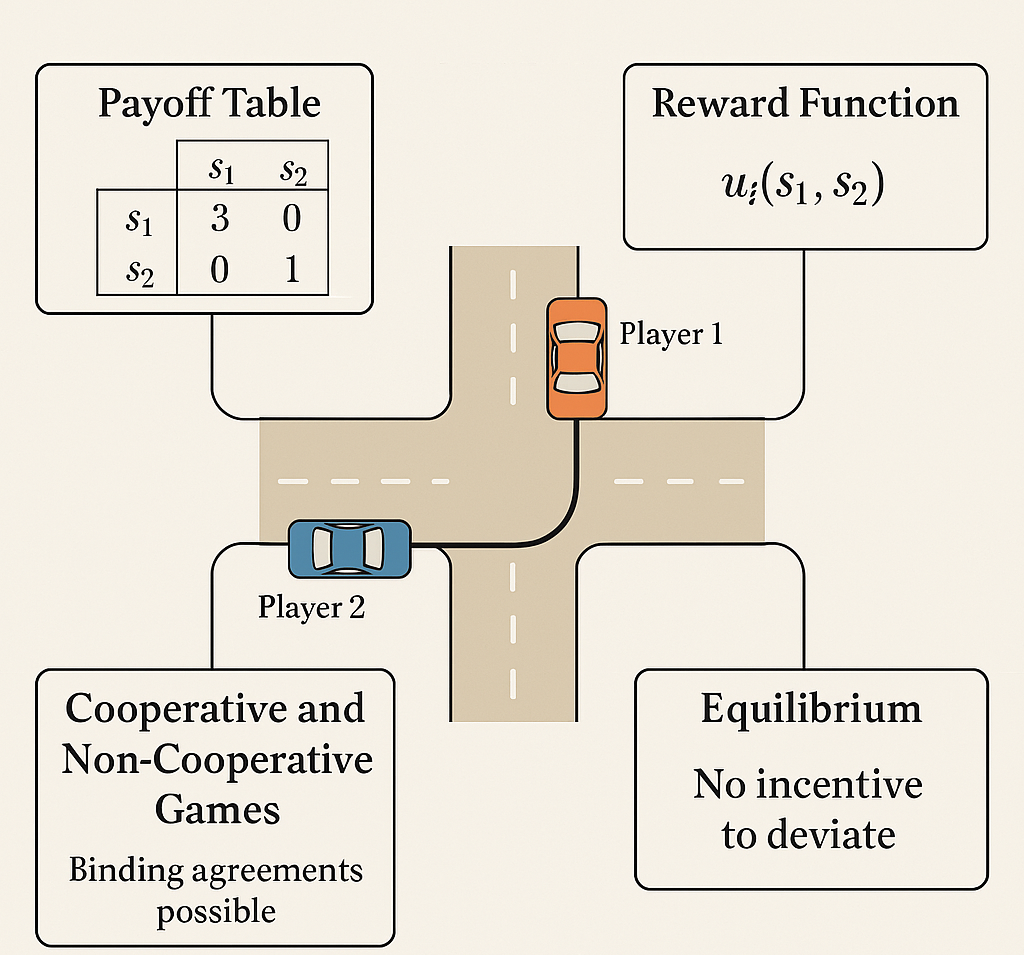}
   \caption{Basic concepts in game theory.}
    \label{fig1}
\end{figure*}
Against this backdrop, game-based approaches offer a compelling alternative for addressing the interactive decision-making challenges inherent in mixed traffic environments~\cite{bacsar1998dynamic,8678699}. By leveraging the principles of game theory, these methods provide a structured framework to model and predict the strategic interactions between AVs and HDVs~\cite{kita2002game,liu2025data}. Game-based strategies can account for equilibrium points and variations in driving styles, thereby offering a more nuanced understanding of the driving dynamics in complex environments~\cite{yu2018human}. The game theory methods not only help to overcome many of the shortcomings observed in traditional approaches but also provide a versatile framework capable of adapting to a range of traffic scenarios. In recent time, Mean-Field-Game guided approaches have been developed for the interactive driving, considering different driving styles~\cite{zheng2025mean,zheng2025enhanced}

It is important to note that the tasks and challenges faced by AVs vary considerably across different driving scenarios. Whether navigating high-speed highways~\cite{claussmann2019review}, managing the complexities of on-ramp merging~\cite{milanes2010automated}, negotiating roundabouts~\cite{zhang2025evaluation}, handling the unpredictability of unsignalized intersections~\cite{xu2019cooperative}, and participating in the autonomous car racing~\cite{betz2022autonomous}, each scenario presents unique demands on interactive decision-making processes. The additional complexity introduced by environments such as autonomous racing further underscores the need for robust, adaptable algorithms. Despite the promise shown by game-based methods, there remains a notable gap in the literature, a comprehensive review of recent game-based interactive driving approaches across these varied scenarios, along with a detailed analysis of their underlying mechanisms, is currently lacking.

This paper seeks to fill that gap by providing an extensive review of game-based interactive driving methods applied across different traffic scenarios. Our objective is to systematically analyze the theoretical underpinnings and practical implementations of these methods, critically assess their performance in handling dynamic interactions, and identify both their strengths and limitations. By offering a detailed examination of how game-based strategies can be optimized to achieve equilibrium in interactive driving scenarios, this paper aims to highlight their adapted mechanisms and equilibrium for improved interactive driving among different scenario. Ultimately, this work contributes to the advancement of intelligent transportation systems by charting a path forward for the integration of game-based approaches into scenario-based interactive driving and outlining future research directions in this evolving field.
\begin{figure*}[t]
    \centering
    \includegraphics[width=0.85\linewidth]{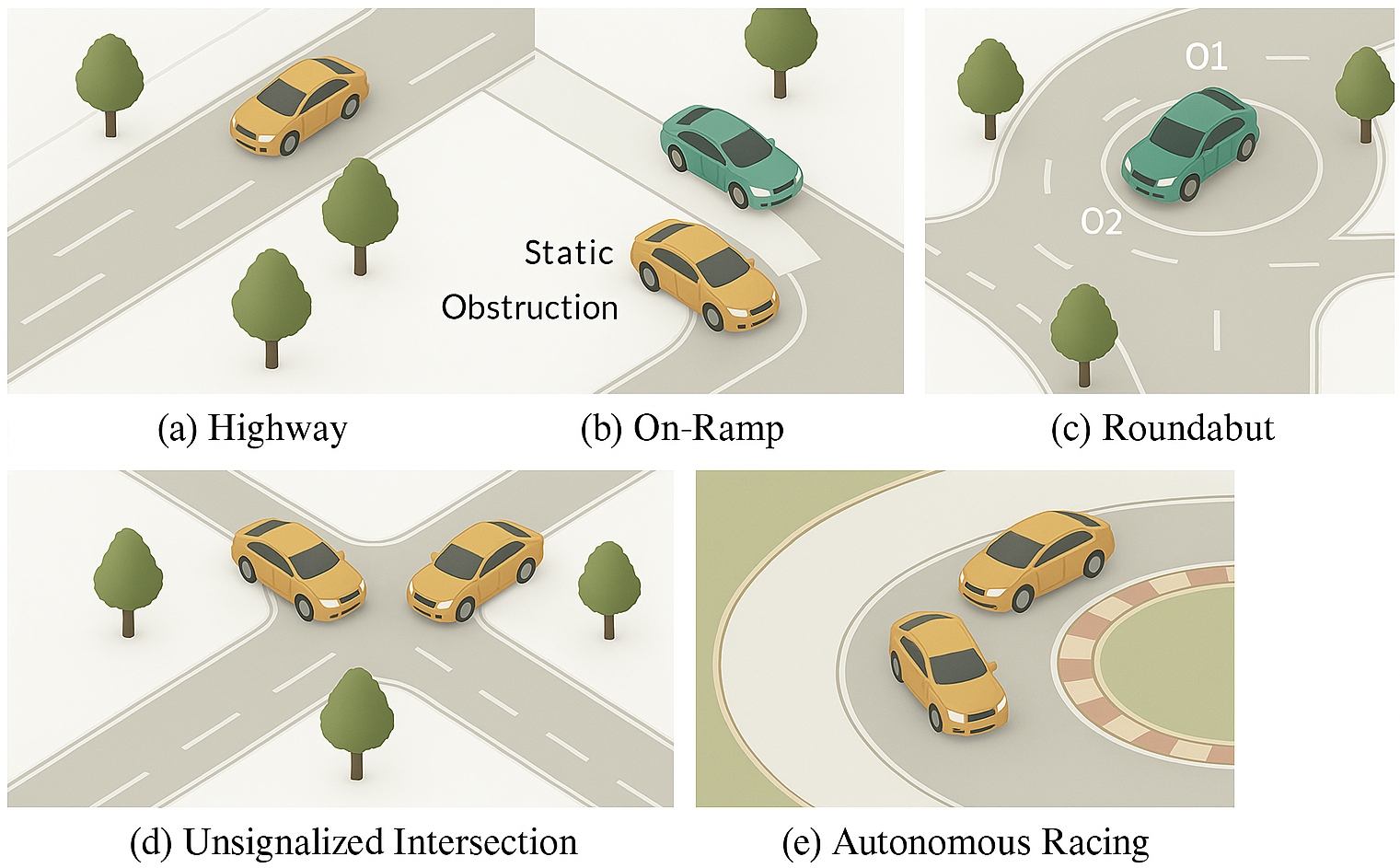}
    \caption{Game-theoretic scenarios for AVs: (a) highway driving, (b) on-ramp merging with blockage, (c) navigating multi-exit roundabouts, (d) negotiating unsignalized intersections, (e) high-speed autonomous racing.}
    \label{fig:game_driving_scenarios}
\end{figure*}

\section{Basic Elements of Game Theory in Driving Scenarios}

Game theory provides a powerful mathematical framework to model strategic interactions among intelligent agents—such as AVs in dynamic traffic environments. In the context of driving, each vehicle selects actions to optimize its own objectives, such as safety, efficiency, comfort, while anticipating the decisions of other vehicles~\cite{chen2023safe}. Key components of game-theoretic modeling in driving include payoff structures, utility functions, equilibrium strategies, and the classification of driving games as either cooperative or non-cooperative~\cite{cominetti2010payoff}.

\subsection{Payoff Matrices in Driving}

A payoff matrix is a structured way to represent the outcomes for different combinations of vehicle strategies. Consider a two-vehicle merging scenario, where vehicle A can either Yield or Accelerate, and vehicle B can either Merge Early or Merge Late. Each action pair results in a payoff depending on safety, time efficiency, and comfort. For vehicle A, the payoff matrix is exampled as:

\[
A = \begin{bmatrix}
2 & 5 \\
1 & 3
\end{bmatrix}
\]

Here, each entry \(a_{ij}\) represents the reward for vehicle A when it selects action \(i\) and vehicle B selects action \(j\). Higher values correspond to more favorable outcomes (e.g., minimal delay and no collision risk).

\subsection{Utility Functions and Objectives}

In driving games, the reward function or utility function \(u_i\) formalizes the individual goals of each vehicle, balancing objectives such as minimizing travel time, avoiding collisions, and maintaining smooth acceleration. For \(N\) vehicles with joint strategy profile \((s_1, \ldots, s_N)\), the utility function for vehicle \(i\) is given by:

\begin{equation}
u_i: S_1 \times \cdots \times S_N \rightarrow \mathbb{R}
\end{equation}

For example, in highway lane-changing, a utility function may penalize sudden deceleration, short following distance, or high risk of collision, while rewarding successful overtaking and speed maintenance.

\subsection{Dominant Strategies and Equilibria}
A strategy is dominant if it yields the highest reward regardless of others’ choices. However, dominant strategies are rare in realistic traffic scenarios. Instead, driving games often analyze Nash equilibrium~\cite{kreps1989nash}, where no vehicle can improve its utility by unilaterally changing its strategy.

Formally, a joint strategy profile \((s_1^*, \ldots, s_N^*)\) is a Nash equilibrium if for every vehicle \(i\),
\begin{equation}
u_i(s_1^*, \ldots, s_i^*, \ldots, s_N^*) \geq u_i(s_1^*, \ldots, s_i, \ldots, s_N^*), \quad \forall s_i \in S_i
\end{equation}

In a roundabout negotiation scenario, a Nash equilibrium corresponds to a mutually acceptable set of decisions where no vehicle benefits from deviating independently.

\subsection{Cooperative vs. Non-Cooperative Driving Games}
In cooperative driving games, vehicles communicate or coordinate their actions to optimize shared outcomes \cite{mariani2021coordination}, such as platooning or intersection crossing via vehicle-to-vehicle (V2V) communication. Solutions may involve shared utility maximization, with methods like bargaining or coalition formation.

In contrast, non-cooperative driving games assume vehicles act independently and may not trust or share intentions with others~\cite{bacsar1998dynamic}, such as AVs interacting with HDVs. These games are often solved using decentralized equilibrium concepts, such as Nash or Stackelberg equilibrium.
Fig.~\ref{fig1} visualizes the key game-theoretic components and their relevance to driving environments.

\section{Game-Theoretic Road Scenarios and Driving Tasks}
This section presents five real-world autonomous driving scenarios, including highways, on-ramping merging, roundabouts, unsignalized intersections, and autonomous racing, as illustrated in Fig.~\ref{fig:game_driving_scenarios}. Each scenario involves complex interactions with HDVs or other AV, modeled as multi-agent games where each vehicle acts to maximize its own driving objectives. Game theory provides a structured framework to reason about strategic behavior, conflict, cooperation, and decision-making under uncertainty in each driving environment.

\subsection{Highways}
Highways are structured, multi-lane environments engineered for uninterrupted high-speed travel, typically involving limited entry/exit points and clearly marked lanes. While highways are generally viewed as predictable, they present nuanced game-theoretic challenges when AVs must continuously reason about surrounding HDVs. Strategic interactions in this setting commonly include car-following games, where the AV determines its acceleration relative to the leading vehicle, and lane-changing games, where it must evaluate whether shifting lanes improves long-term utility.

As illustrated in Fig.~\ref{fig:game_driving_scenarios}(a), the AV drives in the center lane of a three-lane highway surrounded by HDVs that may cruise steadily, brake suddenly, or perform unanticipated lane changes. The AV must decide whether to follow, change lanes, or adjust speed, all while predicting likely responses from adjacent vehicles. The payoff structure typically involves a trade-off between forward progress, collision avoidance, and comfort. The challenge lies in computing robust strategies that maximize expected reward while managing multi-agent uncertainty and partial observability. Here, Nash equilibrium strategies enable the AV to anticipate best responses from nearby drivers, while mixed-strategy reasoning allows for safe decision-making under ambiguity.

\subsection{On-Ramping Merging}
On-ramping merging introduces asymmetry, constrained space, and temporal urgency, making it a classic setting for two-player game-theoretic modeling. The merging AV must negotiate entry into a faster-moving main lane, often interacting with a single nearby HDV that may or may not yield. Both agents are self-interested, and their strategies—such as accelerate, decelerate, or maintain speed—interact non-trivially.

 Fig.~\ref{fig:game_driving_scenarios}(b) shows an AV in a ramp lane approaching a blockage, forcing a merge before the lane ends. The HDVs on the main lanes exhibit varying driving styles, and the AV must estimate their willingness to yield. This creates a decision-making scenario governed by utility estimation under uncertainty. The AV faces a short decision horizon due to the ramp length, while HDV intentions remain partially observable. The driving task is to determine an optimal merging plan that minimizes collision risk while maintaining speed.

In cooperative settings with V2V communication, merging becomes a coalition game with shared utility~\cite{hang2021cooperative}. In non-cooperative conditions, however, the AV must adopt defensive mixed strategies, using probabilistic models to infer likely HDV reactions. Game formulations such as Bayesian games or Stackelberg games are often used to model the interaction.

\subsection{Roundabouts}
Roundabouts are dynamic and decentralized traffic environments where vehicles interact continuously while entering, circulating, and exiting. Unlike signalized intersections, roundabouts require AVs to make nuanced judgments about right-of-way, gap acceptance, and multi-lane positioning. From a game-theoretic perspective, roundabouts introduce a repeated multiplayer setting with time-varying interaction graphs.

In Fig.~\ref{fig:game_driving_scenarios}(c), the AV approaches from EB4 and targets exit O2. It must decide between staying in the outer lane, which offers lower risk but longer travel, or transitioning into the inner lane, which is shorter but riskier due to other vehicles traveling at higher speeds. Strategic decision-making here must account for the predicted actions of HDVs already inside the roundabout, some of whom may behave unpredictably, including accelerating, decelerating, or switching lanes without clear signaling.

Driving tasks in this context involve anticipating route choices of others, minimizing lateral maneuvers, and optimizing exit timing. The AV must continuously update its beliefs about surrounding HDVs and re-evaluate strategies as new agents enter the system. This setting naturally maps to Stackelberg games~\cite{ding2019multivehicle}, where the AV's optimal strategy changes in response to evolving agent configurations.

\subsection{Unsignalized Intersections}
Unsignalized intersections, often encountered in suburban and rural road networks, are multi-agent negotiation zones where centralized traffic control is absent. AVs in such scenarios must reason about driver intentions, predict crossing sequences, and execute safe, socially acceptable maneuvers. Strategic decisions are made in an environment where both right-of-way and movement timing are implicitly communicated.

As shown in Fig.~\ref{fig:game_driving_scenarios}(d), the AV encounters a four-way intersection with three approach lanes per direction. It must choose among left turn, right turn, or going straight, while simultaneously interpreting the likely trajectories and intentions of HDVs arriving from other directions. Compared to roundabouts, the path overlap is more direct and collision-prone, demanding higher temporal precision and more accurate opponent modeling.

Challenges include predicting when another driver will yield, interpreting implicit signals such as creeping forward or vehicle hesitation, and safely entering contested space. These dynamics are well captured by decentralized, non-cooperative games under incomplete information. Bayesian-Nash equilibrium concepts can be used to model mutual belief updates~\cite{jain2022bayesian}, while level-k reasoning or recursive policy inference can approximate bounded rationality in human drivers~\cite{faillo2013roles}.

\subsection{Autonomous Racing}
Autonomous racing introduces a fundamentally different objective function. While safety remains essential, the primary goal is to beat competitors in a high-speed, high-precision environment. AVs must develop both short-term and long-term tactics, making split-second decisions under adversarial pressure.

 Fig.~\ref{fig:game_driving_scenarios}(e) illustrates a racing circuit shared by multiple AVs. Each vehicle attempts to optimize racing lines, acceleration profiles, and overtaking maneuvers while dealing with the dynamic responses of opponents. Strategic actions include blocking, slipstreaming, and cornering aggressiveness. The AV must balance positioning against others’ expected maneuvers and exploit transient advantages.

The racing domain naturally maps to zero-sum or general-sum games, depending on the competition format. Real-time policy adaptation is essential, particularly in multi-lap settings where agents learn each other’s strategies. Game-theoretic models capture continuous-time vehicle dynamics, while Stackelberg games formalize leader-follower behavior during overtaking~\cite{huang2025fair} In this setting, success requires not just safety and coordination, but also deception, risk calibration, and dominance, promoting AV reasoning to its cognitive and computational limits.

\section{Game-based Decision-making on Highways}
Early work applied Stackelberg games to model highway interactions. For example, a hierarchical leader-follower model was used for car-following and lane-changing on highways, and an optimal control formulation for merging was cast as a Stackelberg game. The mixed-motive game theory is used for interactive decision-making in the context of a two-player game involving autonomous vehicles in~\cite{Kim2014}, while the safety and road conditions are considered for both gaming vehicles based on payoff matrices. The lane-changing scenario has been explored between vehicle with and without communication capabilities in \cite{Talebpour2015}. The Nash equilibrium is applied to maximize each vehicle's own profits. However, this approach fails to include analysis of using different driving styles. In~\cite{Wang2015}, cost functions for cooperative and non-cooperative gaming are differently designed, the non-cooperative cost function is formulated as~\cite{Wang2015}:
\begin{equation}
\label{eq:cost_noncoop}
\begin{aligned}
L\bigl(z(t), u(t), t\bigr) \;=\; &\, p_{\mathrm{sfc}}\,\left(\frac{\bigl(x_{\ell} - x - l\bigr)^{2}}{\Delta t^{2}}\right) \\[2ex]
&+\; p_{\mathrm{acc}}\,\bigl(a_{v}^{2}\bigr) \\[2ex]
&+\; p_{\mathrm{dev}}\,\bigl(v - v_{d,n_{\ell}}^{m}\bigr)^{2} \\[2ex]
&+\; p_{\mathrm{lane\,switch}} \,.
\end{aligned}
\end{equation}
where the cost function \(L\bigl(z(t), u(t), t\bigr)\) for the non-cooperative controller is defined as a weighted sum of penalty terms that promote safety, comfort, and efficiency. Specifically, the first term, weighted by \(p_{\mathrm{sfc}}\), penalizes deviations in the spacing between the lead vehicle with position \(x_{\ell}\) and the subject vehicle, with position \(x\) and length \(l\), by considering the squared difference \(\bigl(x_{\ell} - x - l\bigr)^2\) normalized by the square of the time headway \(\Delta t^2\). The second term, scaled by \(p_{\mathrm{acc}}\), penalizes large longitudinal accelerations \(a_{v}\) of the subject vehicle to encourage smooth motion. The third term, with weight \(p_{\mathrm{dev}}\), penalizes deviations of the subject vehicle's speed \(v\) from the desired or maximum allowed speed \(v_{d,n_{\ell}}^{m}\) on lane \(n_{\ell}\). Finally, the term weighted by \(p_{\mathrm{lane\,switch}}\) imposes a penalty on lane-switch events to discourage unnecessary lane changes. The cooperative objective function is formulated as~\cite{Wang2015}:
\begin{equation}
\label{eq:cost_coop}
\begin{aligned}
L\bigl(z(t), u(t), t\bigr)
\;=\;&
\sum_{q=1}^{N}
\Bigl[
p_{\mathrm{safety}}\,C_{\mathrm{safety}}^{(q)} \\[2ex]
&+\; p_{\mathrm{equil}}\,C_{\mathrm{equil}}^{(q)} \\[2ex]
&+\; p_{\mathrm{control}}\,C_{\mathrm{control}}^{(q)} \\[2ex]
&+\; p_{\mathrm{route}}\,C_{\mathrm{route}}^{(q)} \\[2ex]
&+\; p_{\mathrm{lane}}\,C_{\mathrm{lane}}^{(q)} \\[2ex]
&+\; p_{\mathrm{switch}}\,C_{\mathrm{switch}}^{(q)}
\Bigr].
\end{aligned}
\end{equation}

In the cooperative controller cost function, the overall cost is a weighted sum of several key components. These include safety terms that maintain proper headway or spacing, equilibrium terms that encourage smooth traffic states, control effort terms that penalize excessive acceleration or steering, route efficiency terms guiding vehicles toward optimal routes, lane preference costs that capture comfort or regulatory lane usage, and lane-switch penalties to avoid frequent or unsafe lane changes. Each of these components is assigned a weight. Specifically, \(p_{\mathrm{safety}}\) for safety, \(p_{\mathrm{equil}}\) for equilibrium, \(p_{\mathrm{control}}\) for control effort, \(p_{\mathrm{route}}\) for route efficiency, \(p_{\mathrm{lane}}\) for lane preference, and \(p_{\mathrm{switch}}\) for lane-switch penalties, reflecting its relative importance in the overall optimization. Whereas the non-cooperative controller focuses on optimizing each vehicle’s individual cost in isolation, the cooperative approach in \,\eqref{eq:cost_coop} aggregates costs across all vehicles. In doing so, it seeks a joint optimum that benefits the overall system rather than a single vehicle. This results in a higher-dimensional optimization problem but often yields better global performance, improving traffic flow, safety, and efficiency for the fleet as a whole. The trade-off is increased computational complexity and the need for communication or coordination among vehicles, as each vehicle’s decision influences the shared cost function.

A short time-horizon Stackelberg approach can be used in scenarios where the robot plans its actions while anticipating the human’s response in~\cite{Sadigh2018AR}. This is often justified in practice when humans remain relatively conservative, but still adapt to the robot’s behavior. ~\cite{Sadigh2018AR} Let
\begin{equation}
R^0_{u_R,\,u_H} \;=\; \sum_{t=1}^{N} r_t^0\bigl(u^R_t,\;u^H_t\bigr),
\end{equation}
represent the total reward or negative cost from time \(t=1\) to \(t=N\), where \(u^R_t\) and \(u^H_t\) are the robot’s and human’s actions at time \(t\), respectively. In this setup, the human’s best response \(\mathbf{u}_H^*\) is determined by solving a corresponding subproblem. This problem can also be viewed as an underactuated system in which the robot has direct control over its own action \(\mathbf{u}_R\) but only indirect control over the human’s action \(\mathbf{u}_H\). The system dynamics incorporate the human’s response to the robot’s strategy. Evaluating these dynamics requires solving for the human’s best response, \(\mathbf{u}_H^*\). Although it resembles a partially observable or underactuated Markov decision process, the key distinction is that the human’s strategy is endogenously determined by the Stackelberg game.

Given the human’s best response \(\mathbf{u}_H^*\bigl(\ldots\bigr)\), the robot can plan to maximize its own reward function \(R^R\). Formally, the robot’s action \(\mathbf{u}_R^*\) is given by~\cite{Sadigh2018AR}
\begin{equation}
\mathbf{u}_R^*\bigl(x^0,\,\mathbf{u}_R\bigr)
\;=\;
\arg\max_{\mathbf{u}_R}\;
R^R\!\Bigl(\,x^0,\,\mathbf{u}_R,\;\mathbf{u}_H^*\bigl(x^0,\,\mathbf{u}_R\bigr)\Bigr),
\end{equation}
where \(x^0\) is the initial state, \(R^R\) is the robot’s cumulative reward over the short horizon, and \(\mathbf{u}_H^*(x^0,\,\mathbf{u}_R)\) denotes the human’s best response to the robot’s planned actions. In practice, the robot must solve this bilevel problem in real time, which can be computationally challenging. By modeling the human’s response via \(\mathbf{u}_H^*\), the robot can account for how its own actions influence the human’s behavior. This is especially relevant in interactive tasks, such as shared control or human–robot collaboration. However, it also introduces additional complexity, since solving a Stackelberg game typically involves nested optimizations.

The intentions uncertainty of humans are quantified by using the dispositions of each surrounding vehicle in~\cite{Yoo2012Highway}, thus enhancing the fidelity of the simulations.  Building on this, researchers developed non-cooperative game planners for competitive highway scenarios. Game-theoretic model predictive control has been applied to aggressive lane changes and two-car racing, treating vehicles as players in a dynamic game ~\cite{Hang2020Noncoop, Wang2015, Coskun2019, hang2020integrated}. In~\cite{yan2018highway}, a game-theoretic framework is proposed to model the interaction between human drivers in highway driving scenarios. Specifically, a two-person, non-zero-sum, non-cooperative game under complete information is developed to capture the decision-making behavior of drivers. The novel contribution lies in constructing a neural network-based payoff model that accurately describes the rewards of each driver, which is then rigorously calibrated using real traffic data via an improved Gaussian Particle Swarm Optimization (GPSO) method~\cite{higashi2003particle}. This approach not only builds upon previous successes in artificial intelligence for car-following and lane-changing but also provides a robust mechanism to simulate interactive driver behavior under realistic conditions. In~\cite{Hang2020Noncoop}, a hybrid game mode, including Nash and Stackelberg, are ued for a three-lane highway. The Nash game is organized as~\cite{Hang2020Noncoop}:
\begin{equation}
\begin{aligned}
(a_{x,v}^{\mathrm{EC}*}, \sigma^*) &= \arg\min_{a_{x,v}^{\mathrm{EC}},\,\sigma,\,a_{x,v+\sigma}^{\mathrm{AC}}} J^{\mathrm{EC}}(a_{x,v}^{\mathrm{EC}}, \sigma, a_{x,v+\sigma}^{\mathrm{AC}}) \\
a_{x,v+\sigma}^{\mathrm{AC}*} &= \arg\min_{a_{x,v+\sigma}^{\mathrm{AC}}} J^{\mathrm{AC}}(a_{x,v}^{\mathrm{EC}}, \sigma, a_{x,v+\sigma}^{\mathrm{AC}}) \\
\text{s.t.}\quad &\sigma \in \{-1,0,1\},\\[1mm]
&a_{x,v}^{\mathrm{EC}} \in \bigl[a_{x,v}^{\mathrm{min}},\, a_{x,v}^{\mathrm{max}}\bigr],\\[1mm]
&a_{x,v+\sigma}^{\mathrm{AC}} \in \bigl[a_{x,v+\sigma}^{\mathrm{min}},\, a_{x,v+\sigma}^{\mathrm{max}}\bigr],\\[1mm]
&v_{x,l}^{\mathrm{EC}} \in \bigl[v_{x,l}^{\mathrm{min}},\, v_{x,l}^{\mathrm{max}}\bigr],\\[1mm]
&v_{x,l+\sigma}^{\mathrm{AC}} \in \bigl[v_{x,l+\sigma}^{\mathrm{min}},\, v_{x,l+\sigma}^{\mathrm{max}}\bigr]
\end{aligned}
\end{equation}
where \(a_{x,v}^{\mathrm{EC}*}\) denotes the optimal longitudinal acceleration for the Ego Vehicle (\(\mathrm{EC}\)), and \(a_{x,v+\sigma}^{\mathrm{AC}*}\) represents the optimal longitudinal acceleration for the Adjacent Vehicle (\(\mathrm{AC}\)). The variable \(\sigma^*\) is the optimal lane-change decision, where \(\sigma = -1\) indicates a lane change to the left, \(\sigma = 0\) implies remaining in the current lane, and \(\sigma = 1\) signifies a lane change to the right. Here, \(J^{\mathrm{EC}}\) and \(J^{\mathrm{AC}}\) are the objective functions for the Ego and Adjacent Vehicles, respectively. This optimization process efficiently captures the interplay between dynamic decisions by integrating both lane-change and acceleration variables, thereby providing a robust framework for minimizing the overall cost functions under given constraints. An evaluation of this process indicates that, while it effectively identifies strategic advantages—particularly through the leader's decision precedence—it also introduces computational complexity that may require advanced numerical methods or approximations for real-time application. The stkelberg game is formulated as~\cite{Hang2020Noncoop}:

\begin{equation}
\begin{aligned}
\bigl(a_{x,v}^{\mathrm{EC}*}, \sigma^*\bigr) 
&= \arg\min_{\substack{a_{x,v}^{\mathrm{EC}},\,\sigma,\\ a_{x,v+\sigma}^{\mathrm{AC}}}}
J^{\mathrm{EC}}\bigl(a_{x,v}^{\mathrm{EC}},\,\sigma,\,a_{x,v+\sigma}^{\mathrm{AC}}\bigr),
\\
\gamma^2\bigl(a_{x,v}^{\mathrm{EC}}, \sigma\bigr) 
&= \Bigl\{\, z \in \Phi^2 : 
   a_{x,v+\sigma}^{\mathrm{AC}}\bigl(a_{x,v}^{\mathrm{EC}},\,\sigma,\,z\bigr) \,\le\, \Phi^2 
   \\
&\text{s.t.}\quad \sigma \,\in\, \{-1,0,1\},\\
&\quad a_{x,v}^{\mathrm{EC}} \,\in\, 
   \bigl[a_{x,v}^{\mathrm{min}},\, a_{x,v}^{\mathrm{max}}\bigr],\\
&\quad a_{x,v+\sigma}^{\mathrm{AC}} \,\in\, 
   \bigl[a_{x,v+\sigma}^{\mathrm{min}},\, a_{x,v+\sigma}^{\mathrm{max}}\bigr],\\
&\quad v_{x,l}^{\mathrm{EC}} \,\in\, 
   \bigl[v_{x,l}^{\mathrm{min}},\, v_{x,l}^{\mathrm{max}}\bigr],\\
&\quad v_{x,l+\sigma}^{\mathrm{AC}} \,\in\, 
   \bigl[v_{x,l+\sigma}^{\mathrm{min}},\, v_{x,l+\sigma}^{\mathrm{max}}\bigr].
\end{aligned}
\end{equation}

\noindent
where \(\gamma^2(\cdot)\) captures additional conditions or constraints,such as safety margins that relate to \(\mathrm{AC}\)’s response. All accelerations and velocities must lie within their respective minimum and maximum bounds. 
This formulation illustrates how \(\mathrm{AC}\)’s behavior affects \(\mathrm{EC}\)’s decision-making through a Stackelberg framework.

In~\cite{Coskun2019}, the game theory is integrated with the Markov decision process, forming a Markov game (MG). The MG is formulated as~\cite{Coskun2019}:

\begin{equation}
\begin{aligned}
\varphi^{\mathrm{SV}*}\bigl(s(k)\bigr),\; \varphi^{\mathrm{TV}*}\bigl(s(k)\bigr)
=\, & \underset{a^{\mathrm{SV}}(k),\,a^{\mathrm{TV}}(k)}{\arg\max} \\[3ex]
& \sum_{s(k)} \Bigl[ 
R^{p}\Bigl(s(k+1)\,\Big|\, s(k),\\[3ex]
& \,a^{\mathrm{SV}}(k),\,a^{\mathrm{TV}}(k)\Bigr) \\[3ex]
& \quad +\, T\Bigl(s(k+1)\,\Big|\, s(k),\\[3ex]
&\,a^{\mathrm{SV}}(k),\,a^{\mathrm{TV}}(k)\Bigr)
\,V^{p}\Bigl(s(k)\Bigr)
\Bigr].
\end{aligned}
\end{equation}

\begin{equation}
\begin{aligned}
\varphi^{\mathrm{SV}*}\bigl(s(k)\bigr),\; \varphi^{\mathrm{TV}*}\bigl(s(k)\bigr)
=\, & \underset{a^{\mathrm{SV}}(k),\,a^{\mathrm{TV}}(k)}{\arg\max} \\[3ex]
& \Biggl[ \min_{a^{\mathrm{TV}}(k)} \, 
\sum_{s(k)} R^{p}\Bigl(s(k+1)\,\Big|\,\\[3ex] & s(k),\,a^{\mathrm{SV}}(k),\,a^{\mathrm{TV}}(k)\Bigr)
\Biggr] \\[3ex]
& +\, \sum_{s(k)} T\Bigl(s(k+1)\,\Big|\, s(k),\\[3ex]
& \,a^{\mathrm{SV}}(k),\,a^{\mathrm{TV}}(k)\Bigr)
\,V^{p}\Bigl(s(k)\Bigr).
\end{aligned}
\end{equation}

\begin{equation}
\text{where}
\end{equation}

\begin{equation}
\varphi^\mathrm{SV^*}(s(k)), \varphi^\mathrm{TV^*}(s(k)) = (u_x^\mathrm{SV^*}(k), lc^\mathrm{SV^*}(k), u_x^\mathrm{TV^*}(k))
\end{equation}

\begin{equation}
\varphi^\mathrm{SV}(s(k)), \varphi^\mathrm{TV}(s(k)) = (u_x^\mathrm{SV}(k), lc^\mathrm{SV}(k), u_x^\mathrm{TV}(k)),
\tag{7}
\end{equation}

with Markov-perfect equilibrium

\begin{equation}
\begin{aligned}
V^{p}\Bigl(s(k+1)\,\Big|\, & u_{x}^{\mathrm{SV}*}(k),\; lc_{x}^{\mathrm{SV}*}(k),\; u_{x}^{\mathrm{TV}*}(k)\Bigr) \\[2ex]
\;\ge\; \\[2ex]
V^{p}\Bigl(s(k+1)\,\Big|\, & u_{x}^{\mathrm{SV}}(k),\; lc_{x}^{\mathrm{SV}}(k),\; u_{x}^{\mathrm{TV}}(k)\Bigr), \\[2ex]
& \forall\, u_{x}^{\mathrm{SV}}(k),\; \forall\, u_{x}^{\mathrm{TV}}(k).
\end{aligned}
\end{equation}

\begin{equation}
\begin{aligned}
V^{p}\Bigl(s(k+1)\,\Big|\, & u_{x}^{\mathrm{SV}*}(k),\; lc_{x}^{\mathrm{SV}*}(k),\; u_{x}^{\mathrm{TV}*}(k)\Bigr) \\[2ex]
\;\le\; \\[2ex]
V^{p}\Bigl(s(k+1)\,\Big|\, & u_{x}^{\mathrm{SV}}(k),\; lc_{x}^{\mathrm{SV}}(k),\; u_{x}^{\mathrm{TV}}(k)\Bigr), \\[2ex]
& \forall\, u_{x}^{\mathrm{SV}}(k),\; \forall\, u_{x}^{\mathrm{TV}}(k).
\end{aligned}
\end{equation}

\begin{equation}
\begin{aligned}
V^{p}\Bigl(s(k+1)\,\Big|\, & u_{x}^{\mathrm{SV}*}(k),\; lc_{x}^{\mathrm{SV}*}(k),\; u_{x}^{\mathrm{TV}*}(k)\Bigr) \\[2ex]
\;\ge\; \\[2ex]
V^{p}\Bigl(s(k+1)\,\Big|\, & u_{x}^{\mathrm{SV}}(k),\; lc_{x}^{\mathrm{SV}}(k),\; u_{x}^{\mathrm{TV}}(k)\Bigr), \\[2ex]
& \forall\, u_{x}^{\mathrm{SV}}(k),\; \forall\, u_{x}^{\mathrm{TV}}(k).
\end{aligned}
\end{equation}

\begin{equation}
\begin{aligned}
V^{p}\Bigl(s(k+1)\,\Big|\, & u_{x}^{\mathrm{SV}*}(k),\; lc_{x}^{\mathrm{SV}*}(k),\; u_{x}^{\mathrm{TV}*}(k)\Bigr) \\[2ex]
\;\ge\; \\[2ex]
V^{p}\Bigl(s(k+1)\,\Big|\, & u_{x}^{\mathrm{SV}}(k),\; lc_{x}^{\mathrm{SV}}(k),\; u_{x}^{\mathrm{TV}}(k)\Bigr), \\[2ex]
& \forall\, u_{x}^{\mathrm{SV}}(k),\; \forall\, u_{x}^{\mathrm{TV}}(k).
\end{aligned}
\end{equation}

\begin{equation}
\begin{aligned}
V^{p}\Bigl(s(k+1)\,\Big|\, & u_{x}^{\mathrm{SV}*}(k),\; lc_{x}^{\mathrm{SV}*}(k),\; u_{x}^{\mathrm{TV}*}(k)\Bigr) \\[2ex]
\;\ge\; \\[2ex]
V^{p}\Bigl(s(k+1)\,\Big|\, & u_{x}^{\mathrm{SV}}(k),\; lc_{x}^{\mathrm{SV}}(k),\; u_{x}^{\mathrm{TV}}(k)\Bigr), \\[2ex]
& \forall\, u_{x}^{\mathrm{SV}}(k),\; \forall\, u_{x}^{\mathrm{TV}}(k).
\end{aligned}
\end{equation}

\begin{equation}
\begin{aligned}
V^{p}\Bigl(s(k+1)\,\Big|\, & u_{x}^{\mathrm{SV}*}(k),\; lc_{x}^{\mathrm{SV}*}(k),\; u_{x}^{\mathrm{TV}*}(k)\Bigr) \\[2ex]
\;\ge\; \\[2ex]
V^{p}\Bigl(s(k+1)\,\Big|\, & u_{x}^{\mathrm{SV}}(k),\; lc_{x}^{\mathrm{SV}}(k),\; u_{x}^{\mathrm{TV}}(k)\Bigr), \\[2ex]
& \forall\, u_{x}^{\mathrm{SV}}(k),\; \forall\, u_{x}^{\mathrm{TV}}(k).
\end{aligned}
\end{equation}

In above MG, each player's decision at time \(s(k)\) depends solely on the current state rather than on the entire history. The state \(s(k)\) encapsulates all necessary information for decision-making and evolves according to a Markov process, so that the next state \(s(k+1)\) is determined exclusively by the current state and the players' actions. Key variables in this framework include \(s(k)\), which denotes the state of the system; \(V^p(s(k))\), the value function representing the expected cumulative reward from state \(s(k)\) onward when equilibrium strategies are employed; \(u_{x}^{\mathrm{SV}}(k)\) and \(u_{x}^{\mathrm{TV}}(k)\), the control inputs for the surrounding vehicle (SV) and target vehicle (TV), respectively; \(lc_{x}^{\mathrm{SV}}(k)\), the lane-change decision variable for the SV; and the functions \(R^p\) and \(T\), which represent the immediate reward and the state transition, respectively. The equilibrium conditions ensure that when players adopt their equilibrium actions \(u_{x}^{\mathrm{SV}*}(k)\), \(lc_{x}^{\mathrm{SV}*}(k)\), and \(u_{x}^{\mathrm{TV}*}(k)\), the value function evaluated at the resulting state satisfies 
\[
\begin{aligned}
V^{p}\Bigl(s(k+1)\,\Big|\,u_{x}^{\mathrm{SV}*}(k),\,lc_{x}^{\mathrm{SV}*}(k),\,u_{x}^{\mathrm{TV}*}(k)\Bigr) \\[2ex] \ge V^{p}\Bigl(s(k+1)\,\Big|\,u_{x}^{\mathrm{SV}}(k),\,lc_{x}^{\mathrm{SV}}(k),\,u_{x}^{\mathrm{TV}}(k)\Bigr)
\end{aligned}
\]
for all feasible deviations \(u_{x}^{\mathrm{SV}}(k)\) and \(u_{x}^{\mathrm{TV}}(k)\). This inequality guarantees that no player can improve their expected cumulative reward by unilaterally deviating from the equilibrium strategy, thereby establishing a Markov-perfect equilibrium. Overall, the integration of the Markov framework with these equilibrium conditions ensures that each player's strategy is optimal with respect to both immediate and future rewards, which is especially beneficial in dynamic driving. In~\cite{hang2020integrated}, the Stackelberg equilibrium is embedded with the potential filed to conduct safe lane-changing and overtaking. The AV first selects a proper lane-changing time point under Stackelberg equilibrium and then uses APF-based model predictive control (MPC)~\cite{shang2023emergency,liu2025aco} to make safe lane-changing. The Stackelberg is formulated as~\cite{hang2020integrated}:
\begin{equation}
\label{eq:21}
(a_h^*,\,\alpha_h^*)
\;=\;
\arg\min_{a_h,\,\alpha_h}
\Bigl[
h_{x}^{v}
\bigl(
a_h,\;\alpha_h,\;
a_o^*(a_h,\alpha_h),\;\alpha_o^*(a_h,\alpha_h)
\bigr)
\Bigr],
\end{equation}

\begin{equation}
\label{eq:22}
\bigl(a_o^*(a_h,\alpha_h),\;\alpha_o^*(a_h,\alpha_h)\bigr)
\;=\;
\arg\min_{a_o,\,\alpha_o}
\Bigl[
h_{x}^{v}
\bigl(
a_h,\;\alpha_h,\;a_o,\;\alpha_o
\bigr)
\Bigr],
\end{equation}

\noindent
where \(\alpha_o^*(a_h,\alpha_h)\) denotes the obstacle vehicle's decision about lane selection, and \(a_o^*(a_h,\alpha_h)\) denotes the obstacle vehicle's decision about longitudinal acceleration. Here, \(h_{x}^{v}(\cdot)\) represents the cost (or objective) function of the host vehicle, which depends on both the host vehicle’s decisions \(\bigl(a_h,\alpha_h\bigr)\) and the obstacle vehicle’s responses \(\bigl(a_o,\alpha_o\bigr)\). The presented Stackelberg game process is particularly effective for scenarios such as autonomous lane-change decision-making because it leverages a hierarchical, leader–follower structure where the host anticipates the optimal response of the obstacle vehicle. This anticipatory capability allows the host to proactively minimize its cost function by accounting for the follower’s best response, thus leading to more strategic and potentially safer maneuvers compared to simultaneous decision-making frameworks like Nash equilibrium. In contrast, non-cooperative or Nash-based approaches require each vehicle to decide independently without explicit prediction of the other’s reaction, which might result in suboptimal outcomes or even safety risks in highly dynamic environments. However, the Stackelberg framework also introduces additional computational complexity and modeling challenges, as it necessitates solving a bilevel optimization problem in real time. Overall, while the Stackelberg game can offer improved performance in scenarios where strategic foresight is critical, its practical implementation can balance computational efficiency and model accuracy to fully outperform other game-theoretic approaches in dynamic, real-world applications.

Social payoff and potential games have been used to produce safer and more human-like maneuvers on highways~\cite{Schwarting2019, Liu2024PotentialGame}. In~\cite{Schwarting2019}. The Social Value Orientation (SVO)~\cite{murphy2011measuring} is integrated into a noncooperative dynamic game framework, where agents are modeled as utility-maximizing decision makers. The utility function for an ego agent in a two-agent game is defined as~\cite{Schwarting2019}
\begin{equation}
\label{eq:utility}
g_1 = \cos(\varphi_1)\, r_1(\cdot) + \sin(\varphi_1)\, r_2(\cdot),
\end{equation}
where \(r_1\) and \(r_2\) represent the "reward to self" and "reward to other," respectively, and \(\varphi_1\) characterizes the ego agent's SVO angular preference. In this framework, the agent's SVO determines the weighting between self-oriented and other-oriented outcomes. Specifically, agents can be categorized according to their \(\varphi\) values as follows:
\begin{itemize}
    \item \textbf{Altruistic} (\( \varphi \approx \frac{\pi}{2} \)): maximizing the other party's reward.
    \item \textbf{Prosocial} (\( \varphi \approx \frac{\pi}{4} \)): balancing self and other rewards to benefit the group.
    \item \textbf{Individualistic/Egoistic} (\( \varphi \approx 0 \)): maximizing only their own reward.
    \item \textbf{Competitive} (\( \varphi \approx -\frac{\pi}{4} \)): maximizing relative gains over others.
\end{itemize}

The multi-agent dynamic system evolves according to the discrete-time dynamics~\cite{Schwarting2019}
\begin{equation}
\mathbf{x}^{k+1} = \mathcal{F}(\mathbf{x}^{k}, \mathbf{u}^{k}), \quad \text{subject to } c_i(\cdot) \leq 0,
\end{equation}
where \(\mathbf{x}^{k}\) is the state vector, \(\mathbf{u}^{k}\) is the control input, and \(\mathcal{F}(\cdot)\) represents the state transition function. Each agent determines its optimal control policy by solving the following constrained optimization problem:
\begin{equation}
\mathbf{u}_i^*(\mathbf{x}_i^0, \varphi_i) = \arg \max_{\mathbf{u}_i} \; G_i(\mathbf{x}_i^0, \mathbf{u}_i, \mathbf{u}_{-i}, \varphi_i),
\end{equation}
where the total utility over a time horizon \(N\) is given by
\begin{equation}
G_i(\mathbf{x}_i^0, \mathbf{u}, \varphi_i) = \sum_{k=0}^{N-1} g_i(\mathbf{x}^k, \mathbf{u}^k, \varphi_i) + g_i^N(\mathbf{x}^N, \varphi_i).
\end{equation}

The SVO-based game formulation provides a significant improvement over purely individualistic or competitive game models by explicitly incorporating variability in social preferences. By weighting self-oriented and other-oriented rewards according to each agent's SVO angle (\(\varphi\)), this approach produces more realistic and socially compliant interactions. For instance, prosocial agents (\( \varphi \approx \frac{\pi}{4} \)) balance their own rewards with those of others, leading to cooperative behavior that can enhance overall system performance, such as smoother traffic flow or more efficient maneuvering in autonomous driving scenarios. In contrast, traditional models that ignore social considerations often fail to capture the nuances of human-like decision making. However, the SVO-based model also introduces increased computational complexity and challenges in parameter estimation, as accurately inferring each agent's SVO angle from observed behavior can be demanding in real-time applications. Thus, while the SVO-based game approach effectively bridges the gap between individual optimality and social compliance, its practical deployment must carefully balance computational efficiency, model fidelity, and the robustness of parameter estimation.

Others incorporate learning to estimate or adapt to human drivers’ aggressiveness or preferences. For instance, autonomous vehicles can learn lane-changing payoffs online. In \cite{gong2022modeling}, The key innovation of the game-theoretic component in this study is its formulation of the lane-changing decision process as an interactive game among vehicles. By explicitly modeling the strategic interactions between connected and autonomous vehicles (CAVs) and HDVs under foggy conditions, the approach allows CAVs to evaluate and compare the payoffs of different lane-changing maneuvers in real time. This method captures both cooperative and competitive elements of vehicle behavior, enabling a more nuanced and effective decision-making process that improves traffic flow efficiency compared to traditional lane-changing rules. Ultimately, this game-theoretic framework provides a robust foundation for optimizing vehicle interactions in complex, mixed-traffic scenarios under adverse weather conditions.

In~\cite{Lopez2022}, an innovative planning framework is presented for autonomous vehicles that coordinate with human drivers by leveraging a Stackelberg game formulation combined with active information gathering over the human's internal state. A central component of the framework is the use of safety metrics such as Time-to-Collision (TTC)~\cite{vogel2003comparison,sidaway1996time}, which provides a real-time measure of collision risk and enables proactive safety interventions. In the proposed approach, the autonomous vehicle acts as the leader that optimizes its strategy by anticipating the human driver’s best response—effectively incorporating interactive effects on human actions into its decision-making process. Additional mechanisms include the integration of active information gathering, which reduces uncertainty about the human’s intentions and internal state in real time, and the design of cost functions that balance the objectives of performance, safety, and social coordination. These cost functions enable the vehicle to plan not only to minimize its own costs but also to influence the human driver's behavior in a beneficial way. By combining game-theoretic reasoning with robust safety metrics and active sensing, the framework achieves more adaptive, informed, and safer decision-making in mixed-autonomy environments, ultimately demonstrating significant improvements over traditional methods that treat human behavior as exogenous. The Mont Carlo Tree Search (MCTS) and K-level gaming are combined in~\cite{karimi2023deep}, while the MCTS is charged for selecting proper motions and the K-level gaming is for simulating the real-world driving. In~\cite{huang2024highway}, a novel hierarchical framework is proposed for mandatory lane-changing that integrates game theory with data-driven techniques. A key innovation is the introduction of the Relative Driving Style (RDS) metric to capture differences in driving behaviors among vehicles. By leveraging RDS, the authors develop a game selection mechanism that dynamically chooses between static and dynamic game models, ensuring that the interaction between vehicles, modeled as a two-player non-zero-sum, non-cooperative game under complete information, is accurately reflected. The cooperative game approach, which uses Nash equilibrium as the solution concept, effectively simulates the conflict and cooperation between the EV and SVs. Additionally, the integration of offline data processing techniques, such as Bayesian inference and Nash equilibrium-based data imputation, enhances the realism of the model by grounding it in naturalistic driving data. 

Altruistically combinations via multi-agent reinforcement learning (MARL) built on game-theoretic insights have been explored recently ~\cite{Jain2024Survey, Di2021Survey}. These highway-focused studies demonstrate that game-theoretic decision algorithms can improve realism and safety in mixed traffic by anticipating and influencing human driver actions. In highway driving and lane-changing, MARL agents infused with game objectives exhibit more realistic interactive tactics, such as courtesy or selective aggression~\cite{Toghi2022Altruism, Nishi2019MergingRL}. In~\cite{zhou2024highway}, this paper introduces a comprehensive decision-making framework for autonomous driving in merging scenarios that combines data-driven simulation with game-theoretic models. By leveraging inverse reinforcement learning (IRL) on naturalistic driving data, the authors derive reward functions that capture diverse driving characteristics through a deep neural network, with a Nash Equilibrium game layer modeling interactive behaviors between vehicles. Furthermore, a novel temporal-spatial attention-based deep Q-network (TSA-DQN) is developed to simulate vehicle interactions under mixed decision levels, where level‑k game theory is employed to reflect the differences in driver rationality and cognitive processing. This integration of DRL with hierarchical game-theoretic reasoning not only facilitates the simulation of realistic and anthropomorphic driving behaviors but also significantly improves key performance metrics, such as success rate, efficiency, and safety, compared to prior approaches. Overall, the work represents an important step toward more robust and interactive autonomous vehicle planning and evaluation. 

\begin{table*}[t]
\centering
\caption{Summary of Game-theoretic Decision-making Methods on Highways}
\label{tab:game_highway}
\renewcommand{\arraystretch}{1.3}
\begin{tabular}{p{2.8cm} p{3.2cm} p{4.8cm} p{5.8cm}}
\toprule
\textbf{Reference} & \textbf{Scenario / Game Type} & \textbf{Methodology / Innovation} & \textbf{Main Contribution / Limitation} \\
\midrule
Kim (2014) & Two-player mixed-motive game & Payoff matrices with safety and road conditions & Captures interactive AV decision-making; lacks consideration of diverse driving styles \\
Talebpour (2015) & Lane-changing with/without communication & Nash equilibrium & Maximizes individual profit; ignores driving style heterogeneity \\
Wang (2015) & Cooperative vs. non-cooperative lane-changing & Cost functions with safety, comfort, efficiency vs. joint optimization & Highlights trade-off between global traffic efficiency and computation/communication burden \\
Sadigh (2018) & Robot–human short horizon interaction & Stackelberg bilevel optimization & Anticipates human response; computationally intensive for real-time \\
Yoo (2012) & Human intention uncertainty & Dispositions for surrounding vehicles & Improves fidelity of simulation by modeling uncertainty \\
Yan (2018) & Human–human highway driving & Non-cooperative game + NN payoff + GPSO calibration & Data-driven payoff modeling; robust calibration with traffic data \\
Hang (2020) & 3-lane highway & Hybrid Nash + Stackelberg games & Joint lane-change + acceleration optimization; increased complexity \\
Coskun (2019) & Highway interaction & Markov game (MG) & Merges game theory with MDP; ensures Markov-perfect equilibrium \\
Hang (2020, integrated) & Lane-change/overtaking & Stackelberg equilibrium + artificial potential field MPC & Safe maneuvering via anticipatory planning; bilevel optimization costly \\
Schwarting (2019) & Human-like highway maneuvers & Social Value Orientation (SVO) in dynamic games & Captures altruistic/prosocial vs. egoistic/competitive behaviors; challenges in real-time estimation \\
Gong (2022) & Lane-changing under foggy conditions & Interactive game for CAV–HDV & Real-time payoff evaluation; balances cooperative/competitive behavior \\
Lopez (2022) & Human–AV coordination & Stackelberg + active information gathering + TTC safety metric & Incorporates safety and social coordination; reduces uncertainty of human intent \\
Karimi (2023) & Highway driving & Monte Carlo Tree Search (MCTS) + K-level reasoning & Combines search with bounded-rationality gaming; computationally heavy \\
Huang (2024) & Mandatory lane-changing & Relative Driving Style (RDS) + dynamic/static game selection & Captures heterogeneity; Bayesian inference for data imputation; grounded in naturalistic data \\
Toghi (2022), Nishi (2019) & MARL for highway & Altruism/courtesy strategies in MARL & More realistic AV–HDV interaction; influenced by game objectives \\
Zhou (2024) & Merging scenarios & IRL-based reward learning + Nash game + TSA-DQN + Level-$k$ reasoning & Integrates DRL with game theory; improves success, efficiency, and safety \\
\bottomrule
\end{tabular}
\end{table*}

\section{Game-based Decision-making on On-ramping Merging}
Merging at highway entrances poses a challenge requiring negotiation between entering and mainline vehicles. Some works leverage drivers’ social value orientations or interactive rewards to adjust the AV’s strategy, enabling cooperative behaviors like courteous merging and yielding. For example, ~\cite{Toghi2022Altruism} treats mixed‐autonomy driving as a partially‐observable stochastic game (POSG) in which multiple agents with different social preferences must learn to coordinate. Each agent observes only partial information about the environment and optimizes its own reward function while interacting with other agents that are simultaneously learning. The authors propose a MARL framework that leverages a decentralized social reward signal, allowing agents to develop altruistic or cooperative strategies. By adopting a semi‐sequential training procedure and a specialized network architecture, the approach enables agents to learn from experience without requiring an explicit behavior model of the other agents. Experimental evidence suggests that when agents choose social preferences appropriately, ranging from altruistic to selfish—overall traffic safety and flow can improve. Hence, this work demonstrates how a POSG‐based MARL formulation, coupled with a suitable reward design and training methodology, yields more adaptive and beneficial group behavior in mixed‐autonomy driving scenarios. 

Game-theoretic control strategies have been proposed to handle on-ramp merging by modeling the interaction between the merging vehicle and mainline vehicles as a game. A recent study uses a Stackelberg differential game model in a model-predictive control framework, allowing an autonomous car to infer the merging strategy  and generate a safe trajectory accordingly ~\cite{Wei2022Merging}. The aggressiveness are differentiated based on the pay-offs in \cite{Yoo2013Merging} to simulate the real-world driving. Another approach explicitly implements a Stackelberg game for dense traffic merging, where the merging AV acts as leader and the main-lane vehicle as follower, achieving efficient merges in simulation ~\cite{Ji2021Stackelberg}. Beyond Stackelberg models, researchers have explored Nash-equilibrium-based merging decisions. For example, merging behavior in congested traffic has been modeled as a non-cooperative game where each driver’s strategy is at equilibrium ~\cite{Wang2021Ramp}. A novel data-driven simulation framework for multi-vehicle traffic in merging scenarios is presented in~\cite{li2024merging}. The approach uniquely combines IRL with game theory to determine reward function weights from natural driving trajectory data, thereby capturing diverse driving characteristics. A deep neural network (DNN)~\cite{sze2017efficient} is employed to approximate a nonlinear reward function, with a Nash Equilibrium game layer modeling the interplay between cooperative and competitive behaviors among vehicles. In addition, a custom DQN is developed to simulate human-like driving strategies for vehicles on the main road. The simulation results, when compared to natural driving data, demonstrate that the proposed framework can generate realistic, anthropomorphic driving behaviors across various conditions. Overall, these innovations contribute to a dynamic, interactive driving environment that significantly enhances autonomous vehicle testing and evaluation. 

Game-theoretic trajectory planning accounting for interaction forces has improved merge success rates under various driving style assumptions ~\cite{Liu2022ForcedMerge}. In \cite{chen2023game}, the issue of mandatory lane-changing on on-ramps is addressed by highlighting that while existing studies have focused on the modeling of drivers' bounded rationality has been largely overlooked. In practice, drivers do not always form correct expectations about others' decisions; their judgments, influenced by past experience and environmental factors, lead them to make suboptimal choices under bounded rationality. To overcome this limitation, the paper extends the game-based framwwork by employing a Quantum Reaction Equilibrium (QRE) solution to characterize bounded rationality. This model accounts for judgment errors and other unobservable inaccuracies by assuming that, although drivers hold correct average beliefs and tend to choose the strategy with the highest utility, they nevertheless make probabilistic errors. By integrating the QRE into the lane-changing decision-making framework, the proposed method provides a more realistic representation of merging interactions. The improved representation effectively captures both the average rational behavior and the inherent variability of human decision-making. Ultimately, this approach yields a robust and nuanced framework for predicting lane-changing behaviors in complex, mixed-traffic scenarios. Combining game theory with learning, some works train policies that implicitly capture game-theoretic behaviors. A multi-policy reinforcement learning approach was used to learn merging maneuvers that balance aggression and courtesy, yielding decisions comparable to a game-theoretic solution ~\cite{Nishi2019MergingRL}. To handle more complex merging interactions, such as with multiple vehicles, hierarchical reasoning frameworks have been proposed. For instance, an unprotected left-turn merge can be formulated as a game where an autonomous left-turning vehicle predicts oncoming traffic’s responses and optimizes its turn timing ~\cite{Rahmati2017LeftTurn}. A general game-theoretic human–vehicle interaction model has been used to simulate merging scenarios for testing AV decision algorithms ~\cite{Rahmati2020HumanVeh}. IRL has also been used in conjunction with game theory: for example, to infer the hidden “utilities” drivers optimize in merging games, and then design the autonomous policy accordingly ~\cite{Li2024IRLMerge}. These studies show that game-theoretic merging strategies can improve safety and efficiency at highway on-ramps by explicitly modeling the “merge or yield” negotiation.
\begin{table*}[t]
\centering
\caption{Game-theoretic and Learning-based Methods for On-ramp Merging}
\label{tab:onramp_merging}
\renewcommand{\arraystretch}{1.2}
\begin{tabular}{p{3.4cm} p{2.8cm} p{3.1cm} p{7.3cm}}
\toprule
\textbf{Reference} & \textbf{Setting} & \textbf{Game / RL Type} & \textbf{Methodology / Key Idea \& Notes} \\
\midrule
Toghi et al. (2022)~\cite{Toghi2022Altruism} & Mixed-autonomy, partial observability & POSG + MARL, decentralized social reward & Semi-sequential training; agents learn altruistic$\leftrightarrow$selfish policies via social reward; improves safety/flow when social prefs tuned. \\
Wei et al. (2022)~\cite{Wei2022Merging} & On-ramp merging & Stackelberg differential game + MPC & Leader–follower inference of merging strategy; safe trajectory generation; bilevel nature raises real-time complexity. \\
Yoo et al. (2013)~\cite{Yoo2013Merging} & Merging with style heterogeneity & Payoff-based aggressiveness & Differentiates aggressiveness via payoffs to emulate real driving styles. \\
Ji et al. (2021)~\cite{Ji2021Stackelberg} & Dense-traffic merging & Stackelberg (merging AV leader) & Anticipatory merging versus mainline follower; efficient merges in sim; needs reliable follower-response models. \\
Wang et al. (2021)~\cite{Wang2021Ramp} & Congested merging & Non-cooperative Nash & Drivers choose equilibrium strategies; captures strategic contention; may miss courtesy unless encoded. \\
Li et al. (2024)~\cite{li2024merging} & Multi-vehicle merging sim & IRL + Nash game + DNN reward + custom DQN & Learn non-linear reward from naturalistic data; Nash layer models coop/comp interplay; DQN for main-road behavior; realistic, human-like maneuvers. \\
Liu et al. (2022)~\cite{Liu2022ForcedMerge} & Forced/mandatory merges & Interactive trajectory planning & Accounts for interaction forces, improves success under diverse styles. \\
Chen et al. (2023)~\cite{chen2023game} & Mandatory on-ramp lane change & Bounded rationality via QRE & Extends game framework with QRE to model probabilistic errors/judgment noise; more realistic variability than fully rational models. \\
Nishi et al. (2019)~\cite{Nishi2019MergingRL} & Learning merges (balance aggression/courtesy) & Multi-policy RL & Policies implicitly capture game-like behavior; comparable to explicit game solutions. \\
Rahmati et al. (2017)~\cite{Rahmati2017LeftTurn} & Unprotected left-turn merge & Predict–respond game & Leader predicts oncoming responses; optimizes turn timing; hierarchical reasoning. \\
Rahmati \& colleagues (2020)~\cite{Rahmati2020HumanVeh} & Human–vehicle interaction sim & General game-theoretic model & Testbed for AV decision evaluation in merging; configurable interaction rules. \\
Li et al. (2024, IRL)~\cite{Li2024IRLMerge} & Merging games with hidden utilities & IRL + game-theoretic policy design & Infer driver utilities from trajectories, then design AV policy accordingly; aligns AV with human incentives. \\
\bottomrule
\end{tabular}
\end{table*}

\section{Game-based Decision-making on Roundabouts}
At roundabouts, multiple vehicles must cooperatively negotiate entry without explicit signals. Priority-based heuristics can be suboptimal; thus game theory has been invoked to coordinate entering and circulating vehicles. One of the foundational game-theoretic models for vehicle interactions is introduced in~\cite{Li2018TCST}. While their focus was not exclusively on roundabouts, they modeled a generic unsignalized intersection conflict as a non-cooperative game between two drivers. The type of game is a repeated static game: at each time step each driver chooses an acceleration action, and payoffs are defined to encode preferences for safety and progress. The equilibrium applied was a Nash equilibrium at each decision instance, assuming each player best-responds to the other's action. 
specifically addressed a roundabout entry scenario with an autonomous ego vehicle and a human-driven vehicle \cite{Tian2018Roundabout}. They formulated a dynamic game over the roundabout merging process, while the ego vehicle approaches the yield line and the other vehicle circulates in the roundabout. It is assumed that each player optimizes a discounted cumulative reward over a time horizon. They used a Nash equilibrium solution concept in the dynamic game, following the framework of backward induction to compute equilibrium policies.

Mathematically,  ~\cite{Tian2018Roundabout} defined a stage reward function $r_i(t)$ for each vehicle $i$ at time step $t$ that includes multiple features: proximity to collision, progress along the roundabout, etc. For example, one term $\phi_{\text{collision}}(t)$ is an indicator (or penalty) for collision status. The total reward for a horizon of $H$ steps is $R_i = \sum_{t=0}^{H} \beta^t\, r_i(t)$ with discount factor $\beta < 1$. Each vehicle aims to maximize its own $R_i$ by choosing an acceleration profile. The equilibrium condition is that the pair of control sequences $(u_1^*(\cdot), u_2^*(\cdot))$ is such that neither vehicle can improve $R_i$ by unilateral deviation. The solution is computed via dynamic programming on a discretized state-space. In implementation, the ego vehicle uses the estimated aggressiveness of the other driver to select the appropriate reward weights in the game model. Simulations in a roundabout merger showed that the adaptive game-theoretic controller yields safe merges and adjusts behavior appropriately when faced with cautious or aggressive drivers, respectively.
An extended game-theoretic driving models by incorporating the concept of level-k game and Bayesian belief updates \cite{Li2019CDC}. While their test scenario roundabouts as, where drivers may make decisions based on their assumptions about others' rationality level. In the proposed model, a level-0 driver is one who follows a fixed heuristic, a level-1 driver best-responds assuming others are level-0, a level-2 driver best-responds assuming others are level-1, and so on. In practice, human drivers rarely exceed level 2 or 3 in strategic reasoning. \cite{Li2019CDC} formulated a Bayesian game~\cite{brunner2013connection}: each vehicle $i$ has a type corresponding to its level of reasoning, and each maintains a belief distribution over the other vehicle's type. As observations are made based on the other driver's acceleration patterns, the ego vehicle updates its belief using Bayesian inference.

A somewhat different roundabout scenario is explored in~\cite{Ding2020}: the problem of mandatory lane-changing within a roundabout when multiple vehicles are present. They formulated this as a Stackelberg game with two leaders and one follower, effectively modeling a coordinated maneuver among multiple intelligent vehicles. In their setup, two vehicles that need to swap lanes act as leaders, announcing their strategies, while a third vehicle reacts as a follower. The equilibrium concept is a Stackelberg equilibrium: the leaders choose strategies anticipating the follower’s best response. This models a hierarchical decision process often natural in lane change: the vehicles initiating the lane change commit to a plan, and the others then yield or adjust accordingly.

The key novelty of ~\cite{Ding2020} was a cooperative game-based lane-change algorithm that ensures safety and efficiency by design. Previous baseline approaches might handle such multi-vehicle lane changes either with heuristic rules or by treating it as a series of pairwise yield games. Here, they integrated the game model with an optimization-based planner to produce feasible trajectories. Essentially, the Stackelberg game provided the high-level decision, and a model predictive controller then generated continuous acceleration profiles to execute that decision without collision. 

Mathematically, consider vehicles $A$ and $B$ needing to perform a lane change in conflict, and vehicle $C$ as a through vehicle in the outer lane. The authors assign $A$ and $B$ as co-leaders and $C$ as follower. The cost functions include terms for lane deviation (to encourage completing the lane change quickly), time, and a very high cost for collision between any two vehicles. In a Stackelberg framework, $A$ and $B$ jointly choose strategies $x_A, x_B$ (for example, accelerate or decelerate to create a gap) to minimize their costs, assuming $C$ will respond with a strategy $x_C^*(x_A,x_B)$ that minimizes $C$'s cost given $A$ and $B$'s actions. The equilibrium conditions are:
\begin{equation}
x_C^*(x_A,x_B) = \arg\min_{x_C} J_C(x_A,x_B,x_C),
\end{equation}
and 
\begin{equation}
(x_A^*,x_B^*) = \arg\min_{x_A,x_B} J_{A+B}(x_A,x_B, x_C^*(x_A,x_B)),
\end{equation}
where $J_{A+B}$ is a combined cost of the leaders (possibly a weighted sum if they have separate objectives). Solving this analytically is complex, so ~\cite{Ding2020} resorted to an iterative algorithm that searches for optimal $x_A, x_B$ while simulating the follower’s best response via an MPC for $C$. They demonstrated in simulation of a dual-lane roundabout that this coordinated approach (a) avoids deadlock where both lane-changing vehicles wait for each other, and (b) avoids collision by ensuring the follower’s reaction is always accounted for. The results showed improved travel times and comfort compared to a non-game-theoretic baseline where each vehicle made decisions myopically. This work is an example of using Stackelberg game models for collaborative maneuvers in a roundabout, extending beyond the simpler leader-yielder paradigm to multi-vehicle cooperation.

The problem of autonomous decision-making at roundabout-like yield scenarios is formulated from a human-factors perspective \cite{Li2022IMechE}. They employed game theory with a prospect-theoretic payoff design. Prospect theory, originating from behavioral economics, accounts for the observation that humans perceive gains and losses nonlinearly and relative to a reference point. In this work, the authors crafted the utility functions for the game not as the objective risk or time, but as a subjective utility that mimics human judgement. Specifically, they map the objective collision probability to a subjective ``loss'' value using a nonlinear function, and similarly map time or speed to a subjective ``gain''. By doing so, an autonomous vehicle's decision-making becomes more aligned with human-like tendencies. For example, being more averse to increasing an already small gap, which humans find disproportionately risky.

The type of game model used by \cite{Li2022IMechE} is a two-player dynamic game similar to a Nash game, but with the twist that each player's payoff is given by a prospect-theory-based value function. They consider a scenario of one autonomous vehicle and one human driver approaching a conflict point. Each has two choices: go or yield, at each time step until one passes first. The equilibrium concept is Nash equilibrium in these choices, the equilibrium better reflects what two human drivers might do. The contribution here is introducing fuzzy risk weighting and stochastic driver models into the game formulation, which was a departure from the perfectly rational player assumption. This is novel compared to baseline game models that typically assume linear utility or expected values of outcomes.

For mathematical highlights, \cite{Li2022IMechE} define a subjective utility function $U_i$ for vehicle $i$ as:
\begin{equation} 
U_i = w_{\text{time}}\cdot V\Big( - t_i \Big) + w_{\text{safety}}\cdot V\Big( -p(\text{collision}_i) \Big),
\end{equation}
where $V(\cdot)$ is a value function from prospect theory and $p(\text{collision}_i)$ is the estimated probability of collision for vehicle $i$ if both proceed. The value function $V(x)$ has the typical S-shape: it is concave for gains and convex for losses, with a steeper slope for losses. In effect, a small probability of a severe collision is perceived by the algorithm as a disproportionately large loss, thus encouraging yielding. The game equilibrium is found by comparing the utilities: if $U_{\text{go},A} > U_{\text{yield},A}$ and simultaneously $U_{\text{go},B} > U_{\text{yield},B}$, then both would go, potentially causing a conflict. The system is designed such that in equilibrium one will choose to yield if the situation is unsafe, breaking the symmetry in a human-like manner. The authors validated their model by comparing the autonomous vehicle's decisions against human driver decisions in similar simulations, finding a closer match than using a conventional Nash game with linear costs. This work is significant as it integrates psychological realism into game-theoretic driving which is important for mixed traffic scenarios.

A comprehensive decision-making framework is presented for CAVs at unsignalized roundabouts, explicitly modeling personalized driving styles and comparing different game-theoretic approaches in \cite{Hang2022}. They formulated the interaction among multiple vehicles as a non-zero-sum game and examined two equilibrium concepts: a standard Nash equilibrium and a Stackelberg equilibrium. In addition, they considered a cooperative grand coalition scenario as a theoretical benchmark. Each CAV in their framework has a utility function that incorporates safety, efficiency, and comfort. Crucially, the relative weighting of these factors differs per vehicle to represent different driving styles of aggressive, normal, and cautious. 

The contributions of \cite{Hang2022} are multi-fold. First, it integrates a motion prediction module into the game: using a potential field and MPC-based predictor, each CAV forecasts the short-term trajectory of others, rather than assuming static positions. This prediction feeds into the utility calculation. Second, it demonstrates a hardware-in-the-loop (HIL) real-time implementation, lending credibility to the approach's practicality. Third, by comparing Nash vs. Stackelberg vs. cooperative outcomes, \cite{Hang2022} highlight the advantages of a Stackelberg game for roundabouts: it yielded significantly lower costs for individual vehicles than the Nash simultaneous move solution, because the sequential decision structure helps avoid deadlocks and unnecessary waiting. At the same time, the cooperative game gave the best system efficiency but might not respect individual preferences. Thus, their work balances individual rationality and system optimality by advocating a Stackelberg approach where, say, the first vehicle to arrive acts as leader and others follow in order, achieving near-cooperative efficiency but in a decentralized, incentive-compatible way.

Another study applied game theory for vehicle-to-vehicle negotiation in a roundabout, demonstrating via simulation that basic game strategies can avoid deadlock and collisions ~\cite{Banjanovic2016}. Game-theoretic planning has also been combined with optimal control for conflict resolution in roundabouts. A hierarchical planning framework ~\cite{Fisac2019} treats roundabout negotiation as a two-level game: a higher-level game to decide which car yields and a lower-level optimal control to execute the maneuver safely. Human driving data have been used to calibrate such models, ensuring that the autonomous strategy is socially acceptable. In ~\cite{Shu2023HumanInspired}, a game-theoretic intersection handling strategy uses a Nash bargaining solution to mimic human-like courtesy while still merging efficiently. Several works integrate learning or adaptation into game models at roundabouts. For example, a MCTS based planner has been developed for autonomous vehicle motion planning in roundabouts and other multi-vehicle settings ~\cite{Wang2021KBTree}, enabling the AV to simulate different strategies and select an optimal action that accounts for other drivers’ likely responses. Overall, by treating roundabout entry as a game, these approaches achieve smoother merging and reduce stalemate situations compared to popular benchmarks, such as rule-based methods.
\begin{table*}[t]
\centering
\caption{Game-theoretic and Learning-based Decision-making Methods for Roundabouts}
\label{tab:roundabout_games}
\renewcommand{\arraystretch}{1.2}
\begin{tabular}{p{3.2cm} p{2.8cm} p{3.2cm} p{7.3cm}}
\toprule
\textbf{Reference} & \textbf{Scenario} & \textbf{Game Type} & \textbf{Methodology / Contribution} \\
\midrule
Li et al. (2018)~\cite{Li2018TCST} & Generic unsignalized intersection & Repeated static Nash game & Two drivers repeatedly choose acceleration; payoffs encode safety/progress; Nash equilibrium at each step. \\
Tian et al. (2018)~\cite{Tian2018Roundabout} & Roundabout entry (AV vs HDV) & Dynamic Nash game & Backward induction with discounted rewards; adapts to cautious vs aggressive drivers via reward weighting. \\
Li et al. (2019)~\cite{Li2019CDC} & Roundabout / intersection & Level-$k$ + Bayesian game & Models bounded rationality (level-0,1,2 reasoning); Bayesian belief updates on driver type. \\
Ding et al. (2020)~\cite{Ding2020} & Dual-lane roundabout lane-change & Multi-leader Stackelberg & Two leaders coordinate lane swap, follower reacts; MPC executes trajectories; avoids deadlock/collision. \\
Li et al. (2022)~\cite{Li2022IMechE} & Roundabout-like yield & Prospect-theoretic dynamic game & Subjective utilities (loss/gain asymmetry); more human-like risk aversion; Nash equilibrium with prospect payoffs. \\
Hang et al. (2022)~\cite{Hang2022} & CAVs at unsignalized roundabouts & Non-zero-sum (Nash / Stackelberg / Cooperative) & Personalized driving styles; potential-field prediction + MPC; Stackelberg achieves near-cooperative efficiency; HIL validation. \\
Banjanovic et al. (2016)~\cite{Banjanovic2016} & Vehicle negotiation at roundabout & Basic game strategies & Demonstrated avoidance of deadlock and collisions through simple game interactions. \\
Fisac et al. (2019)~\cite{Fisac2019} & Conflict resolution in roundabouts & Hierarchical game + optimal control & High-level game decides yield/go; low-level optimal control executes safely; calibrated with human data. \\
Shu et al. (2023)~\cite{Shu2023HumanInspired} & Intersection / roundabout & Nash bargaining & Mimics human-like courtesy; balances safety and efficiency. \\
Wang et al. (2021)~\cite{Wang2021KBTree} & Roundabout + multi-vehicle & MCTS-based planning with game rollouts & Simulates opponent strategies; selects optimal AV action; reduces stalemate vs rule-based methods. \\
\bottomrule
\end{tabular}
\end{table*}

\section{Game-based Decision-making on Unsignalized Intersections}
Unsignalized intersections require vehicles to mutually decide who goes first, often modeled as a priority conflict game. Game-theoretic models of uncontrolled intersections have proliferated in recent years. A seminal approach models two vehicles approaching an intersection as players in a chicken-game: each can either yield or go, and the outcomes (potential collision or delay) define a game matrix. This was extended to multi-vehicle interactions using differential games and solved for Nash equilibria ~\cite{Li2020Unsignalized, Tian2020Unsignalized}. These models help formalize intuitive rules as equilibrium strategies emerging from drivers’ utility trade-offs. Follow-up work has tackled more complex intersection scenarios using Bayesian game frameworks and mixed-strategy solutions. An intersection negotiation algorithm in ~\cite{Nan2022Nash} predicts each driver’s intention (aggressive vs. conservative) and computes a mixed-strategy Nash equilibrium for crossing order, improving safety at uncontrolled junctions. Other researchers proposed switchable game modes at intersections: vehicles switch between cooperative and competitive game formulations depending on context, such as all players yielding or one asserting priority, to better model human variations ~\cite{Jia2023Switch}. Reinforcement learning has also been combined with game-theoretic intersection handling. Some methods use game-theory to design the reward structure or initial policy for multi-agent RL, ensuring convergence to socially-desirable behaviors. For example, an AV trained with a game-theoretic reward will learn to sometimes “wave through” another vehicle akin to yielding if it leads to higher overall efficiency ~\cite{Zhao2021Yield}. A recent survey ~\cite{Qin2024Review} summarizes how game theory has been applied to intersection management, including both analytical game models and learning-based approaches. Many studies report that incorporating concepts like reciprocity, altruism, and negotiation from game theory leads to more robust and human-compatible behavior at intersections without signals. Monte Carlo and Tree Search methods have been particularly useful under uncertainty at intersections. By simulating many possible action sequences, MCTS-based planners can identify safe intersection-crossing policies that approximate game-theoretic outcomes ~\cite{Wang2021KBTree}. In~\cite{yuan2022}, a novel framework is presented that combines deep reinforcement learning with game-theoretic level-k game to enable adaptive decision-making at unsignalized intersections. By modeling different reasoning levels using techniques such as adapted DQN, the approach improves coordination and safety among interacting vehicles. Moreover, hierarchical games have been used for multi-vehicle intersections: for instance, one layer computes a Nash equilibrium assuming perfect information, and another layer uses Bayesian inference or partially observable game models to handle occlusions and uncertainty about other drivers ~\cite{Negash2023Review}. Such combinations of game theory and search/planning have demonstrated high success rates in complex unsignalized intersection simulations.

Across all scenarios, a notable trend is combining game theory with machine learning to handle the stochastic and diverse nature of human drivers. Several works use MARL with game-theoretic priors, such as learning policies that achieve a Nash equilibrium or refining a game-theoretic strategy via self-play. In unsignalized intersections, deep Q-learning policies~\cite{zhang2017energy} have been improved by incorporating game-theoretic reward structures and attention mechanisms to predict opponents’ moves ~\cite{zhou2024highway}. These learning-enhanced approaches aim to combine the best of both worlds: the rational decision frameworks of classical game theory, and the adaptive, data-driven strengths of modern AI.
\begin{table*}[t]
\centering
\caption{Game-theoretic and Learning-based Decision-making Methods for Unsignalized Intersections}
\label{tab:unsignalized_intersections}
\renewcommand{\arraystretch}{1.2}
\begin{tabular}{p{3.2cm} p{2.8cm} p{3.2cm} p{7.3cm}}
\toprule
\textbf{Reference} & \textbf{Scenario} & \textbf{Game Type} & \textbf{Methodology / Contribution} \\
\midrule
Li et al. (2020), Tian et al. (2020)~\cite{Li2020Unsignalized, Tian2020Unsignalized} & Two-/multi-vehicle intersection & Chicken game, differential game, Nash equilibrium & Yield/go modeled as chicken-game; extended to multi-vehicle via differential games; equilibrium captures intuitive priority rules. \\
Nan et al. (2022)~\cite{Nan2022Nash} & Intersection negotiation & Mixed-strategy Nash & Predicts drivers’ aggressiveness; computes mixed-strategy equilibrium for crossing order; improves safety at junctions. \\
Jia et al. (2023)~\cite{Jia2023Switch} & Context-aware intersections & Switchable cooperative/competitive games & Vehicles adapt game mode depending on context; models human variation in yielding vs asserting priority. \\
Zhao et al. (2021)~\cite{Zhao2021Yield} & Intersection crossing & Game-theoretic reward shaping + RL & AV learns to “wave through” another vehicle when beneficial; improves overall efficiency and safety. \\
Qin et al. (2024)~\cite{Qin2024Review} & Survey of intersection methods & Review (game theory + learning) & Summarizes applications of analytical game models and ML-integrated approaches for intersection management. \\
Wang et al. (2021)~\cite{Wang2021KBTree} & Intersections under uncertainty & MCTS-based planning & Simulates action sequences to approximate game-theoretic outcomes; robust to uncertainty. \\
Yuan et al. (2022)~\cite{yuan2022} & Adaptive decision-making & Level-$k$ game + DRL (DQN) & Models reasoning levels via level-$k$; deep RL improves coordination and safety among interacting vehicles. \\
Negash et al. (2023)~\cite{Negash2023Review} & Multi-vehicle intersections & Hierarchical games + Bayesian/PO games & Two-layer structure: Nash for perfect info, Bayesian inference for uncertainty; handles occlusion and incomplete info. \\
Zhang et al. (2017)~\cite{zhang2017energy}, Zhou et al. (2024)~\cite{zhou2024highway} & Intersection crossing with DRL & Game-theoretic reward + attention mechanisms & DQN policies improved with game-theoretic priors and attention; predicts opponent moves more effectively. \\
\bottomrule
\end{tabular}
\end{table*}

\section{Game-based Decision-making on Autonomous Racing}
The multi-vehicle gaming is achieved by using Nash equilibrium in~\cite{ Wang2019RSS}, while the main tasks for the competitors are remaining within the racing track regions. The objective function is formulated as~\cite{ Wang2019RSS}:
\begin{equation}
\label{eq:best_response}
\begin{aligned}
s_i(\theta_{-i})
&= \arg\min_{\theta_i}\; h_i\bigl(\theta_i,\theta_{-i}\bigr),\\
\text{s.t.}\quad
&g_i\bigl(\theta_i,\theta_{-i}\bigr) \,\le\, 0,\\
&\gamma_i\bigl(\theta_i,\theta_{-i}\bigr) \,\le\, 0,
\end{aligned}
\end{equation}

In this formulation, \(s_i(\theta_{-i})\) is the best-response function for player \(i\), given the strategies \(\theta_{-i}\) of all other players. The function 
\[
h_i(\theta_i,\theta_{-i})
\]
represents a new objective, typically combining the reward of player \(i\) with the negative rewards of other players. This added term captures the competitive aspect: in addition to maximizing its own progress or payoff, player \(i\) is motivated to decrease the payoffs of others by strategically adjusting \(\theta_i\). Despite introducing this new objective, it can be shown that any Nash equilibrium arising from \(h_i(\cdot)\) is also a Nash equilibrium of the original game. Therefore, solving the game with this surrogate objective function still yields valid solutions for the original problem, while making explicit the competitive incentives among the players. The constraints 
\[
g_i(\theta_i,\theta_{-i}) \le 0
\quad\text{and}\quad
\gamma_i(\theta_i,\theta_{-i}) \le 0
\]
capture any physical, safety, or strategic limitations, such as collision avoidance or performance bounds, that must be respected. The design for autonomous racing integrates non-cooperative and cooperative cost functions within a game-theoretic framework that enables vehicles to optimize both individual performance and collective interactions. The non-cooperative component focuses on individual objectives such as safe spacing, controlled acceleration, and adherence to desired speeds, ensuring that each vehicle operates within safe and efficient bounds. In contrast, the cooperative cost function aggregates these objectives across all vehicles, promoting coordinated behavior that can enhance overall race dynamics through tactics like drafting or synchronized overtaking. The best-response formulation further captures the competitive nature of racing by allowing each vehicle to adjust its strategy in response to the anticipated actions of its competitors, thereby balancing aggressive maneuvers with safety considerations. While this approach is theoretically robust and well-suited for dynamic, multi-agent environments, its practical implementation will require careful tuning of weighting parameters and advanced computational techniques in the fast-paced context of autonomous racing.

Some extensions include sensitivity-enhanced iterative best-response planners and control barrier functions to enhance robustness \cite{Notomista2020CBF}.

Stackelberg games assume leader–follower dynamics that are useful in overtaking scenarios \cite{Liniger2019Game,Jung2023gameMPC}. In~\cite{Liniger2019Game}, a novel game-based framework is proposed for interactive trajectory planning in autonomous driving. For online implementation, the proposed  repeats the game in a receding horizon manner, akin to MPC, thereby generating a sequence of coupled games that effectively address higher-dimensional challenges typically intractable via conventional dynamic programming methods. A key innovation is the integration of modified constraints based on viability theory. These constraints not only ensure recursive feasibility with respect to track requirements but also extend this guarantee to the actions of the opponent vehicle, thanks to an exact soft constraint reformulation that maintains feasibility at all times.

In~\cite{Liniger2019Game}, three distinct game formulations are presented for interactive trajectory planning in competitive driving scenarios, each addressing different strategic objectives:

\textbf{Sequential Game:} In the sequential game formulation, the payoff matrices are designed such that the leader (P1) focuses exclusively on maximizing progress while remaining agnostic to collision outcomes. In contrast, the follower (P2) is penalized for collisions. This asymmetric structure aligns with typical racing dynamics, where the leader’s actions are evaluated solely based on progress, whereas the follower must prioritize collision avoidance to achieve a feasible trajectory. The resulting formulation guarantees that even if collisions occur, P1 is unaffected, making it especially suited to environments where a clear leader-follower hierarchy exists.

\textbf{Cooperative Game:} The cooperative game extends collision considerations to both players. Here, both P1 and P2 receive the progress payoff only when their joint trajectories remain feasible—that is, without leaving the track or incurring a collision. By ensuring that collision avoidance is a shared objective, this formulation harmonizes the strategies of both participants. Consequently, the cooperative game fosters a mutual understanding of feasibility, leading to trajectory pairs that are jointly optimized for safety and progress, as evidenced by the modified payoff matrices which penalize any collision event for both players.

\textbf{Blocking Game:} Recognizing that, in racing scenarios, the ultimate goal is not only to progress but also to finish first, the blocking game modifies the payoff structure further. In this formulation, an additional blocking reward parameter \(w\) is introduced to encourage strategies that prevent the opponent from overtaking, even at the expense of some progress. This trade-off is explicitly captured in the modified payoff matrices where a lower progress value for one player can be compensated by a strategic reward when the other player is blocked. Thus, the blocking game successfully captures the dual objectives of advancing forward and maintaining a positional advantage in competitive environments.

Overall, \cite{Liniger2019Game} demonstrates that these game-theoretic formulations, offer unique advantages tailored to specific racing objectives. The sequential game is best suited for scenarios with a clear leader-follower structure, the cooperative game promotes mutual safety through shared collision avoidance, and the blocking game introduces a tactical dimension by incentivizing strategic blocking maneuvers. Together, these formulations provide a comprehensive framework that advances our understanding of interactive decision-making in competitive and safety-critical driving applications.

In~\cite{Jung2023gameMPC}, a data-driven hyperparameter optimization scheme is proposed for the identification of vehicle system model parameters. The proposed approach combines a random search strategy with explore–exploit theory to allocate computational budget efficiently. By defining a budget that determines both the number of sampled configurations and the iterations allocated to each, Hyperband systematically prunes configurations with high evaluation losses while allocating more resources to promising candidates. To further enhance the search and mitigate the risk of converging to local minima, The proposed approach incorporates a Gaussian mutation process, which injects stochastic perturbations into the selected configurations, thereby increasing the likelihood of identifying a solution with a lower evaluation loss. An essential component of the planner is an aggressiveness scaling mechanism that adaptively switches between two levels, \(\alpha_{\text{inactive}}\) and \(\alpha_{\text{active}}\). The parameter \(\alpha\) scales the aggressiveness of the ego agent in the potential game formulation, allowing it to either race conservatively when opponents are distant or assertively when opponents are near. The switching logic is controlled by comparing the average squared distance between the ego agent and the other agents to a predefined threshold \(D\). This process is summarized in Algorithm~\ref{algo:rapid} and can be expressed mathematically as~\cite{Jung2023gameMPC}
\begin{equation}
    \alpha = 
    \begin{cases}
    \alpha_{\text{inactive}}, & \text{if } \displaystyle \sum_{i \in [N]} \Bigl\lVert x^\mathrm{ego} - x^i \Bigr\rVert^2 > (N-1)D,\\[8pt]
    \alpha_{\text{active}},   & \text{otherwise.}
    \end{cases}
    \label{eq:alpha_switch}
\end{equation}
where \(N\) is the total number of agents, and \(x^\mathrm{ego}\) and \(x^i\) denote the states of the ego agent and the \(i\)-th opponent, respectively.

\begin{algorithm}[t]
\caption{Interactive Decision-Making Planner~\cite{Jung2023gameMPC}}
\label{alg:interactive_planner}
\begin{algorithmic}[1]
    \State \textbf{Input:} Sensor data \(s(t)\), current state estimate \(x(t)\), dynamic obstacle set \(\mathcal{O}(t)\), prediction horizon \(T\), vehicle and environment parameters

    \State \textbf{Step 1: State Estimation and Obstacle Prediction}
    \State Process sensor data to compute current state \(x(t)\) and estimate dynamic obstacles.
    \For{each obstacle in \(\mathcal{O}(t)\)}
        \State Predict future trajectory \(\hat{o}(t+\Delta t)\) using motion models.
    \EndFor
    \State Identify critical obstacles and regions of high risk.

    \State \textbf{Step 2: Game Formulation and Payoff Evaluation}
    \State Formulate the interactive game using vehicle dynamics and safety constraints.
    \State Define payoff functions \(\Phi\) based on progress and collision risk.
    \State Construct payoff matrices that encode the strategic interactions between the agent and obstacles.

    \State \textbf{Step 3: Strategy Generation and Selection}
    \For{each candidate strategy \(g \in \mathbf{G}\)}
        \State Generate candidate trajectory in Frenet coordinates \((l(t), d(t))\) using quintic polynomials.
        \State Evaluate the candidate using the payoff function \(\Phi\) and check for smooth transitions.
    \EndFor

    \State \textbf{Return:} Optimal trajectory set \(\mathbf{G}^*\) corresponding to the highest payoff.
\end{algorithmic}
\end{algorithm}

\begin{figure*}[t]
    \centering
    \includegraphics[width=1\linewidth]{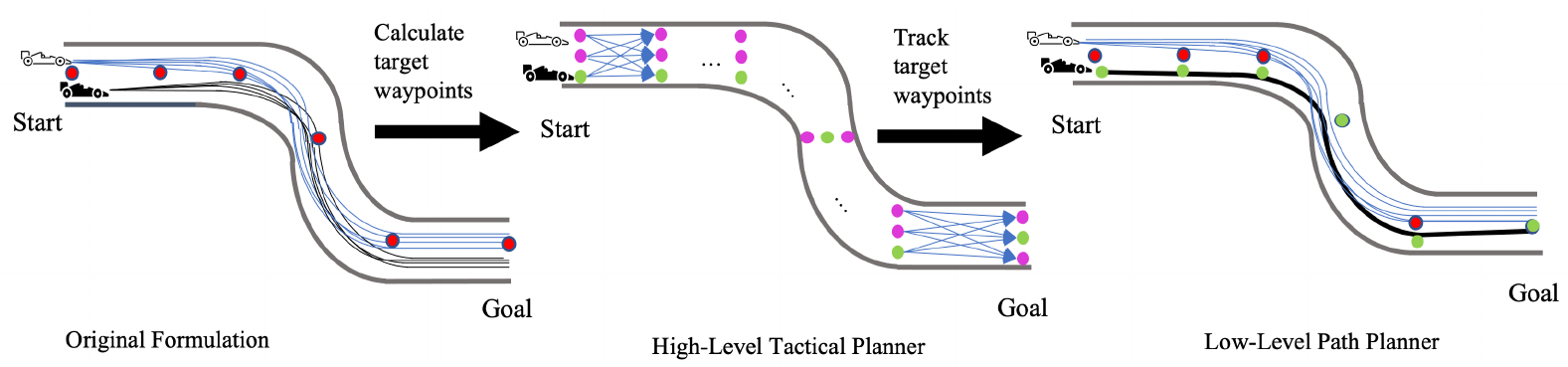}
    \caption{The uncountably infinite trajectories of the general game (left) discretized by the high-level planner (middle). The sequence of target waypoints calculated by the high-level planner (in green) is tracked by the low-level planner (right) and converges to a continuous trajectory (in black).~\cite{Thakkar2024FR}}
    \label{figlh}
\end{figure*}
The proposed hierarchical control scheme in \cite{Thakkar2024FR} effectively addresses the dual challenge of achieving competitive head-to-head performance while rigorously complying with the complex safety and etiquette rules of racing. Unlike traditional methods that simplify these rules to basic collision avoidance, the proposed formulation accommodates discrete constraints, such as limits on lane changes, resulting in a mixed-integer nonlinear optimization that models a realistic head-to-head racing game.

In this architecture, the high-level tactical planner constructs a fully discretized abstraction of the game. This abstraction encodes intricate rules and plans over an extended time horizon by generating a sequence of discrete target waypoints. In parallel, the low-level path planner operates on a simplified version of the original problem, tracking these target waypoints under a reduced set of constraints, based on the assumption that the high-level planner has already accounted for the nuanced rules. Advanced solution methods, such as Monte Carlo tree search, linear-quadratic Nash game formulations, and multi-agent reinforcement learning, are then exploited to solve the simplified games efficiently in real-time.

Fig.~\ref{figlh} illustrates a hierarchical racing planner composed of three stages. On the left, the “Original Formulation” shows the theoretically infinite set of possible paths a vehicle could take from the start to the goal on a winding track. This reflects the continuous, high-complexity nature of the underlying game, where the agent could follow any one of infinitely many trajectories. In the middle, the “High-Level Tactical Planner” is responsible for reducing this vast search space by selecting a series of discrete target waypoints. The colored dots in this illustration represent those waypoints, strategically placed along the track to satisfy both progress and rule-adherence objectives. On the right, the “Low-Level Path Planner” follows these target waypoints to create a smooth, continuous path (shown in black). By focusing on the set of waypoints chosen by the tactical planner, the low-level planner can efficiently and accurately generate a trajectory that integrates adherence to racing constraints, collision avoidance, and performance goals—all while operating in real-time.

Recent studies integrate high-level Monte Carlo tree search for decision-making and generalized Nash equilibrium problems (GNEP) for low-level trajectory optimization, ensuring compliance with fairness and racing regulations \cite{Huang2025sports}.

Hierarchical frameworks combining MARL and game-theoretic principles have emerged \cite{Thakkar2024FR,Jia2023RAPID}. An agent racing algorithm that employs constrained dynamic potential games for multi-agent racing scenarios has been presented in \cite{Jia2023RAPID}. Unlike conventional non-cooperative constrained general-sum dynamic games, the proposed racing game is an instance of a dynamic potential game. In such games, equilibria always exist and can be determined by solving a single constrained optimal control problem.

The key advantage of formulating the problem as a dynamic potential game is that it is generally more efficient to solve the underlying multivariate optimal control problem than to address a set of coupled optimal control problems. The proposed planning algorithm for racing among multiple agents has been compared with state-of-the-art methods and has demonstrated improvements in both computation time and trajectory quality.

Equilibrium computations frequently involve iterative best-response algorithms, sequential quadratic programming (SQP)~\cite{boggs1995sequential}, and iterative linear-quadratic game approximations (iLQGames) in \cite{Rowold2024IV}. Recent advancements include dynamic game sequential quadratic programming (DG-SQP) improving GNE solutions \cite{Zhu2024GNE}:
\begin{equation}
x_{k+1} = x_k - H_k^{-1}\nabla L(x_k,\lambda_k), \quad \lambda_{k+1} = \lambda_k + \alpha_k g(x_{k+1}).
\end{equation}
\begin{table*}[t]
\centering
\caption{Game-theoretic and Learning-based Decision-making Methods for Autonomous Racing}
\label{tab:racing_games}
\renewcommand{\arraystretch}{1.2}
\begin{tabular}{p{3.2cm} p{2.6cm} p{3.2cm} p{7.4cm}}
\toprule
\textbf{Reference} & \textbf{Scenario} & \textbf{Game Type} & \textbf{Methodology / Contribution} \\
\midrule
Wang et al. (2019)~\cite{Wang2019RSS} & Multi-vehicle racing on track & Non-cooperative Nash equilibrium & Best-response optimization under safety/track constraints; objective balances self-progress and opponent payoff reduction; captures competitive racing nature. \\
Notomista et al. (2020)~\cite{Notomista2020CBF} & Robust racing interactions & Iterative best-response + CBF & Sensitivity-enhanced planners with control barrier functions; robustness against disturbances and collisions. \\
Liniger et al. (2019)~\cite{Liniger2019Game} & Competitive racing maneuvers & Sequential / Cooperative / Blocking games & Three formulations: (i) sequential leader–follower; (ii) cooperative joint collision avoidance; (iii) blocking game encouraging tactical blocking. Implemented with receding horizon MPC. \\
Jung et al. (2023)~\cite{Jung2023gameMPC} & Data-driven racing MPC & Stackelberg / Potential game with aggressiveness scaling & Hyperparameter optimization with Hyperband + Gaussian mutation; aggressiveness parameter \(\alpha\) switches based on opponent proximity; interactive MPC with payoff adaptation. \\
Thakkar et al. (2024)~\cite{Thakkar2024FR} & Hierarchical racing with etiquette rules & Mixed-integer hierarchical game + MCTS & High-level tactical planner discretizes rules into waypoints; low-level planner tracks feasible trajectories; integrates Monte Carlo tree search and Nash formulations. \\
Huang et al. (2025)~\cite{Huang2025sports} & Competitive sports-like AV racing & MCTS + GNEP & High-level MCTS for decision-making; generalized Nash equilibrium for low-level trajectory optimization; compliance with fairness and racing regulations. \\
Jia et al. (2023)~\cite{Jia2023RAPID} & Multi-agent racing & Dynamic potential game & Constrained dynamic potential game ensures equilibrium existence; solved as single optimal control problem; improves computation time and trajectory quality. \\
Rowold et al. (2024)~\cite{Rowold2024IV} & Real-time multi-agent racing & iLQGames + SQP & Iterative linear-quadratic approximations for game equilibrium; sequential quadratic programming improves convergence. \\
Zhu et al. (2024)~\cite{Zhu2024GNE} & Multi-agent racing optimization & DG-SQP for GNE & Dynamic game SQP method accelerates convergence to generalized Nash equilibrium in racing contexts. \\
\bottomrule
\end{tabular}
\end{table*}

\section{Conclusion and Analysis}

Game-theoretic decision-making frameworks have become indispensable in modeling the strategic interactions of autonomous vehicles (AVs) in complex driving environments. As autonomous driving systems evolve to navigate increasingly dynamic, mixed-traffic scenarios, the integration of game theory provides a rigorous foundation for handling multi-agent decision-making, negotiation, and social interaction. This section summarizes the current developments in game-based decision-making across five core driving scenarios: highways, on-ramping merging, roundabouts, unsignalized intersections, and autonomous racing. It highlights their distinct features, key challenges, successful models, and outlines promising directions for future research, grounded in the detailed survey presented above.

\subsection{Scenario-Specific Discussion}

On highways, the core strategic challenges involve continuous negotiation over space and velocity through car-following, lane-changing, and gap acceptance. Classical Stackelberg models have effectively captured leader-follower dynamics for longitudinal and lateral motion, enabling AVs to anticipate reactions from HDVs. For example, in \cite{Wang2015}, cooperative and non-cooperative cost formulations allow vehicles to minimize control efforts while maintaining safety and efficiency. Moreover, advanced approaches integrate Social Value Orientation (SVO) into dynamic games to reflect the varying levels of altruism or competitiveness among drivers, as demonstrated in \cite{Schwarting2019}. Recent work introduces bounded rationality via quantal response equilibria and probabilistic strategy errors, such as in \cite{chen2023game}, offering greater realism. Other studies, like \cite{hang2020integrated}, employ hybrid models combining Stackelberg games with artificial potential fields and MPC for safe lane changes. A notable direction is the use of MARL to learn equilibrium strategies that adaptively capture social behavior, such as courtesy or tactical blocking, as illustrated in \cite{gong2022modeling}.

On-ramping merging emphasizes short-horizon, high-stakes interactions between entering vehicles and main-lane traffic. Traditional models, such as \cite{Wei2022Merging}, use Stackelberg differential games embedded in MPC frameworks, enabling AVs to predict merging feasibility based on whether they act as a leader or follower. In \cite{Toghi2022Altruism}, merging is framed as a partially observable stochastic game, with decentralized MARL agents learning to balance aggressiveness and social cooperation. More nuanced driver behavior is modeled in \cite{Yoo2013Merging} and \cite{chen2023game} using data-driven estimation of driver aggressiveness and bounded rationality. Realistic merging strategies have also emerged through inverse reinforcement learning combined with Nash equilibrium reasoning, as shown in \cite{li2024merging}. Notably, \cite{Liu2022ForcedMerge} models forced merges where insufficient space exists, addressing rare but critical safety challenges.

Roundabouts present unique topological and behavioral complexity, with vehicles entering and exiting in continuous flow without explicit signals. The work of \cite{Tian2018Roundabout} formulates dynamic entry games over finite horizons, allowing ego AVs to plan acceleration profiles that anticipate human driver behavior. Level-$k$ games and Bayesian reasoning have been applied in \cite{Li2019CDC} to simulate different cognitive levels, enabling AVs to infer and adapt to varying aggressiveness. Multi-vehicle interactions are modeled through cooperative Stackelberg games, such as in \cite{Ding2020}, which enable co-leaders to coordinate strategies with a third follower vehicle, ensuring efficient and collision-free maneuvers. Prospect theory-based payoffs in \cite{Li2022IMechE} introduce psychological realism by encoding asymmetric risk perception, improving yield-vs-go decisions in ambiguous merges. Studies like \cite{Hang2022} show that Stackelberg-based hierarchical models offer near-cooperative efficiency without requiring full coordination.

At unsignalized intersections, game-based models offer structured ways to resolve ambiguous priority and reduce collision risks. Early studies model conflict resolution using chicken games and Nash equilibria, while more recent work expands into Bayesian games with mixed strategies that adapt to driver personality estimates \cite{Nan2022Nash}. For example, in \cite{yuan2022}, deep reinforcement learning agents with level-$k$ game reasoning improved intersection safety and coordination. Other works, such as \cite{Jia2023Switch}, dynamically switch between cooperative and non-cooperative games based on perceived context, mimicking human reasoning. Prospect-theoretic utilities have also been used to model loss aversion in time-gap acceptance decisions. Real-world scalability is demonstrated through Monte Carlo tree search-based planning with integrated game solvers \cite{Wang2021KBTree}.

Autonomous racing scenarios represent the most competitive and adversarial domain for game-theoretic planning. These settings feature zero-sum or general-sum interactions among vehicles aiming to overtake or block one another at high speeds. Nash equilibrium formulations, such as in \cite{Wang2019RSS}, compute competitive policies subject to safety and progress constraints. In \cite{Liniger2019Game}, sequential, cooperative, and blocking games are defined for various racing goals, including aggressive defense and shared safety. Hierarchical planners, like those in \cite{Thakkar2024FR}, decouple strategic waypoint planning from low-level tracking, enabling safe yet competitive navigation. Dynamic potential games and equilibrium solvers such as DG-SQP in \cite{Zhu2024GNE} allow for scalable multi-agent racing with fairness constraints. Adaptive aggression scaling and hyperparameter optimization, as used in \cite{Jung2023gameMPC}, further customize the AV's behavior based on opponent proximity.

\subsection{Future Research Directions}
Future research in highway scenarios should focus on scaling existing models to handle high-density traffic and heterogeneous vehicle capabilities. Incorporating multi-lane, multi-agent interactions into equilibrium computation requires approximate solvers and decentralized planning. Learning-based policy distillation from real traffic, hierarchical game decompositions, and intent estimation modules can improve response accuracy while maintaining computational feasibility.

For on-ramping merging, promising directions include the integration of V2X communication to coordinate merging strategies, learning-based intent inference for HDVs under occlusion, and robust policy transfer across weather and traffic conditions. Further enhancement can be achieved by simulating population-level driving styles using large-scale naturalistic datasets.

In roundabouts, hybrid planning that combines fast motion prediction, SVO inference, and adaptive game-type switching could improve safety and efficiency. Development of modular planning stacks with plug-and-play game models for different traffic contexts could accelerate deployment. Exploiting human trajectory data via imitation learning to refine reward and belief models is another promising avenue.

At unsignalized intersections, deeper integration of probabilistic perception with hierarchical game models is essential to manage occlusions, partial observability, and rare aggressive behaviors. Innovations in online learning and real-time driver modeling will enable AVs to adapt to changing environments and previously unseen driver types.

Autonomous racing offers a unique sandbox to explore generalization, adversarial tactics, and long-term strategy in constrained high-speed domains. Future work should explore scalable approximations to generalized Nash equilibrium, opponent modeling using meta-learning, and simulation-to-reality transfer for racetrack deployment. Designing interpretable reward shaping and hierarchical controllers will also enhance explainability and trust in AV racing systems.

Across all domains, blending game theory with reinforcement learning, intention inference, and hierarchical control remains the most promising strategy to develop safe, robust, and socially compliant autonomous vehicles operating in diverse and dynamic real-world environments.

\subsection{Functional Outlook and Development Implications}

Based on the literature surveyed, the development of game-based decision-making for autonomous driving holds transformative potential across application domains. In highway driving, game-theoretic policies are no longer just theoretical constructs as they have matured to the point where interpretable models can be directly embedded in planning stacks, ensuring measurable gains in lane-keeping stability, fuel efficiency, and human-compliance. In on-ramping scenarios, functional deployment is already feasible for limited domains using Stackelberg-based MPCs that enable AVs to negotiate merges under time and space constraints. Scenarios like those in \cite{Wei2022Merging} could be implemented in near-term systems equipped with V2X and onboard trajectory prediction.

For roundabouts and unsignalized intersections, game-based systems are progressing toward maturity, especially with the integration of level-$k$ and SVO reasoning into AV planners. These approaches offer scalable alternatives to reactive or rule-based systems, enabling AVs to respond with human-like sensitivity to uncertainty and implicit negotiation. The potential is particularly evident in shared urban spaces, where data-driven calibration of game parameters, such as in \cite{Li2022IMechE} or \cite{yuan2022}, can improve acceptance, comfort, and safety for all road users.

Autonomous racing, while niche, serves as a critical proving ground for high-fidelity strategic reasoning and long-horizon competition. The architectures developed, including hierarchical MCTS planners and DG-SQP equilibrium solvers, offer transferable insights for dense traffic and multi-agent highway coordination. Research in this domain is now poised to inform not only entertainment and robotics competitions but also emergency response and tactical maneuvers in congested urban environments.

Overall, game-theoretic AV decision-making is evolving from descriptive modeling toward functional deployment. What began as simplified matrix games has now grown into dynamic, interpretable, and data-driven multi-agent planning frameworks capable of operating in real-world scenarios. Future AV systems will benefit from incorporating these capabilities, offering more explainable, adaptive, and ethically grounded behavior across a spectrum of mobility applications.

\footnotesize\bibliographystyle{IEEEtran}
\bibliography{IEEEabrv,zq_lib}

\begin{thebibliography}{100}
\providecommand{\url}[1]{#1}
\csname url@samestyle\endcsname
\providecommand{\newblock}{\relax}
\providecommand{\bibinfo}[2]{#2}
\providecommand{\BIBentrySTDinterwordspacing}{\spaceskip=0pt\relax}
\providecommand{\BIBentryALTinterwordstretchfactor}{4}
\providecommand{\BIBentryALTinterwordspacing}{\spaceskip=\fontdimen2\font plus
\BIBentryALTinterwordstretchfactor\fontdimen3\font minus \fontdimen4\font\relax}
\providecommand{\BIBforeignlanguage}[2]{{%
\expandafter\ifx\csname l@#1\endcsname\relax
\typeout{** WARNING: IEEEtran.bst: No hyphenation pattern has been}%
\typeout{** loaded for the language `#1'. Using the pattern for}%
\typeout{** the default language instead.}%
\else
\language=\csname l@#1\endcsname
\fi
#2}}
\providecommand{\BIBdecl}{\relax}
\BIBdecl

\bibitem{muratori2021rise}
M.~Muratori, M.~Alexander, D.~Arent, M.~Bazilian, P.~Cazzola, E.~M. Dede, J.~Farrell, C.~Gearhart, D.~Greene, A.~Jenn \emph{et~al.}, ``The rise of electric vehicles—2020 status and future expectations,'' \emph{Progress in Energy}, vol.~3, no.~2, p. 022002, 2021.

\bibitem{duarte2018impact}
F.~Duarte and C.~Ratti, ``The impact of autonomous vehicles on cities: A review,'' \emph{Journal of Urban Technology}, vol.~25, no.~4, pp. 3--18, 2018.

\bibitem{de2018itssafe}
A.~M. De~Souza, L.~L. Pedrosa, L.~C. Botega, and L.~Villas, ``Itssafe: An intelligent transportation system for improving safety and traffic efficiency,'' in \emph{2018 IEEE 87th Vehicular Technology Conference (VTC Spring)}.\hskip 1em plus 0.5em minus 0.4em\relax Ieee, 2018, pp. 1--7.

\bibitem{jagatheesaperumal2024artificial}
S.~K. Jagatheesaperumal, S.~E. Bibri, J.~Huang, J.~Rajapandian, and B.~Parthiban, ``Artificial intelligence of things for smart cities: advanced solutions for enhancing transportation safety,'' \emph{Computational Urban Science}, vol.~4, no.~1, p.~10, 2024.

\bibitem{10477849}
D.~Liu, L.~T. Yang, R.~Zhao, X.~Deng, C.~Zhu, and Y.~Ruan, ``Multi-tree compact hierarchical tensor recurrent neural networks for intelligent transportation system edge devices,'' \emph{IEEE Transactions on Intelligent Transportation Systems}, vol.~25, no.~8, pp. 8719--8729, 2024.

\bibitem{10932698}
S.~Song and M.~Fan, ``Emergency routing protocol for intelligent transportation systems using iot and generative artificial intelligence,'' \emph{IEEE Transactions on Intelligent Transportation Systems}, pp. 1--12, 2025.

\bibitem{lin2025conflicts}
Z.~Lin, Z.~Tian, J.~Lan, Q.~Zhang, Z.~Ye, H.~Zhuang, and X.~Zhao, ``A conflicts-free, speed-lossless kan-based reinforcement learning decision system for interactive driving in roundabouts,'' \emph{IEEE Transactions on Intelligent Transportation Systems}, 2025.

\bibitem{zuo2025industrial}
Q.~Zuo \emph{et~al.}, ``Industrial internet robot collaboration system and edge computing optimization,'' \emph{arXiv preprint arXiv:2504.02492}, 2025.

\bibitem{fujiang2025ai}
Y.~Fujiang \emph{et~al.}, ``Ai-driven optimization of blockchain scalability, security, and privacy protection,'' \emph{Algorithms}, vol.~18, no.~5, p. 263, 2025.

\bibitem{lin2025safety}
Z.~Lin, J.~Lan, C.~Anagnostopoulos, Z.~Tian, and D.~Flynn, ``Safety-critical multi-agent mcts for mixed traffic coordination at unsignalized intersections,'' \emph{IEEE Transactions on Intelligent Transportation Systems}, 2025.

\bibitem{zuo2025advanced}
Q.~Zuo, C.~Li, Y.~Fan, W.~Kang, T.~Yang, N.~Feng, Z.~Tian, and M.~Yu, ``Advanced multi-modal sensor fusion architectures for robust autonomous driving systems,'' in \emph{2025 IEEE 5th International Conference on Electronic Technology, Communication and Information (ICETCI)}.\hskip 1em plus 0.5em minus 0.4em\relax IEEE, 2025, pp. 1556--1560.

\bibitem{he2025integrated}
Z.~He, R.~Xu, B.~Wang, Q.~Meng, Q.~Tang, L.~Shen, Z.~Tian, and J.~Duan, ``Integrated blockchain and federated learning for robust security in internet of vehicles networks,'' \emph{Symmetry}, vol.~17, no.~7, p. 1168, 2025.

\bibitem{yang2025application}
Q.~Yang \emph{et~al.}, ``The application of intelligent optical sensor networks in industrial automation,'' in \emph{Fifth International Conference on Telecommunications, Optics, and Computer Science (TOCS 2024)}, vol. 13629.\hskip 1em plus 0.5em minus 0.4em\relax SPIE, 2025, pp. 60--67.

\bibitem{ju2025research}
W.~Ju, Y.~Dong, J.~Zhu, Z.~Tian, and Q.~Li, ``Research on real-time vehicle detection and tracking algorithm based on dense optical flow,'' in \emph{Intelligent Transportation and Smart Cities}.\hskip 1em plus 0.5em minus 0.4em\relax IOS Press, 2025, pp. 146--154.

\bibitem{zuo2025intelligent}
H.~Zuo \emph{et~al.}, ``Intelligent road crack detection and analysis based on improved yolov8,'' in \emph{2025 8th International Conference on Advanced Algorithms and Control Engineering (ICAACE)}.\hskip 1em plus 0.5em minus 0.4em\relax IEEE, 2025, pp. 1192--1195.

\bibitem{zhong2025skybound}
J.~Zhong, X.~Fang \emph{et~al.}, ``Skybound magic: Enabling body-only drone piloting through a lightweight vision--pose interaction framework,'' \emph{International Journal of Human--Computer Interaction}, pp. 1--31, 2025.

\bibitem{qin2025ppa}
W.~Qin, J.~Li \emph{et~al.}, ``Ppa-enhanced yolov10: real-time detection of small uavs in complex environments,'' in \emph{Second International Conference on Image Processing and Artificial Intelligence (ICIPAI 2025)}, vol. 13780.\hskip 1em plus 0.5em minus 0.4em\relax SPIE, 2025, pp. 375--381.

\bibitem{10704959}
D.~Chu, C.~Zhao, R.~Wang, Q.~Xiao, W.~Wang, and D.~Cao, ``A survey of multi-vehicle consensus in uncertain networks for autonomous driving,'' \emph{IEEE Transactions on Intelligent Transportation Systems}, vol.~25, no.~12, pp. 19\,319--19\,341, 2024.

\bibitem{xiong2023review}
X.~Xiong, S.~Zhang, and Y.~Chen, ``Review of intelligent vehicle driving risk assessment in multi-vehicle interaction scenarios,'' \emph{World Electric Vehicle Journal}, vol.~14, no.~12, p. 348, 2023.

\bibitem{kyung2010enhancing}
G.~Kyung, M.~A. Nussbaum, and K.~L. Babski-Reeves, ``Enhancing digital driver models: Identification of distinct postural strategies used by drivers,'' \emph{Ergonomics}, vol.~53, no.~3, pp. 375--384, 2010.

\bibitem{moser2010managing}
P.~R. Moser, ``Managing unsafe drivers and their unsafe habits,'' in \emph{ASSE Professional Development Conference and Exposition}.\hskip 1em plus 0.5em minus 0.4em\relax ASSE, 2010, pp. ASSE--10.

\bibitem{salmon2019bad}
P.~M. Salmon, G.~J. Read, V.~Beanland, J.~Thompson, A.~J. Filtness, A.~Hulme, R.~McClure, and I.~Johnston, ``Bad behaviour or societal failure? perceptions of the factors contributing to drivers' engagement in the fatal five driving behaviours,'' \emph{Applied ergonomics}, vol.~74, pp. 162--171, 2019.

\bibitem{BADUE2021113816}
C.~Badue, R.~Guidolini, R.~V. Carneiro, P.~Azevedo, V.~B. Cardoso, A.~Forechi, L.~Jesus, R.~Berriel, T.~M. Paixao, F.~Mutz \emph{et~al.}, ``Self-driving cars: A survey,'' \emph{Expert systems with applications}, vol. 165, p. 113816, 2021.

\bibitem{UKGov2023}
{Department for Transport}, ``Road accidents and safety statistics,'' \url{https://www.gov.uk/government/collections/road-accidents-and-safety-statistics}, 2023, accessed: 2024-04-28.

\bibitem{BASt2020}
\BIBentryALTinterwordspacing
``Verkehrs- und unfalldaten - kurzzusammenstellung der entwicklung in deutschland,'' Bundesanstalt für Straßenwesen (BASt), Bergisch Gladbach, Germany, Tech. Rep., 2020, technical report. [Online]. Available: \url{https://www.bast.de/DE/Publikationen}
\BIBentrySTDinterwordspacing

\bibitem{tao2022advanced}
W.~Tao, M.~Aghaabbasi, M.~Ali, A.~H. Almaliki, R.~Zainol, A.~A. Almaliki, and E.~E. Hussein, ``An advanced machine learning approach to predicting pedestrian fatality caused by road crashes: A step toward sustainable pedestrian safety,'' \emph{Sustainability}, vol.~14, no.~4, p. 2436, 2022.

\bibitem{lin2024dpl}
Z.~Lin, Q.~Zhang, Z.~Tian, P.~Yu, and J.~Lan, ``Dpl-slam: enhancing dynamic point-line slam through dense semantic methods,'' \emph{IEEE Sensors Journal}, vol.~24, no.~9, pp. 14\,596--14\,607, 2024.

\bibitem{lin2025slam2}
Z.~Lin, Q.~Zhang, Z.~Tian, P.~Yu, Z.~Ye, H.~Zhuang, and J.~Lan, ``Slam2: Simultaneous localization and multimode mapping for indoor dynamic environments,'' \emph{Pattern Recognition}, vol. 158, p. 111054, 2025.

\bibitem{lin2024enhanced}
Z.~Lin, Z.~Tian, Q.~Zhang, H.~Zhuang, and J.~Lan, ``Enhanced visual slam for collision-free driving with lightweight autonomous cars,'' \emph{Sensors}, vol.~24, no.~19, p. 6258, 2024.

\bibitem{omeiza2021explanations}
D.~Omeiza, H.~Webb, M.~Jirotka, and L.~Kunze, ``Explanations in autonomous driving: A survey,'' \emph{IEEE Transactions on Intelligent Transportation Systems}, vol.~23, no.~8, pp. 10\,142--10\,162, 2021.

\bibitem{ignatious2022overview}
H.~A. Ignatious, M.~Khan \emph{et~al.}, ``An overview of sensors in autonomous vehicles,'' \emph{Procedia Computer Science}, vol. 198, pp. 736--741, 2022.

\bibitem{perez2011autonomous}
J.~P{\'e}rez, V.~Milan{\'e}s \emph{et~al.}, ``Autonomous driving manoeuvres in urban road traffic environment: a study on roundabouts,'' \emph{IFAC Proceedings Volumes}, vol.~44, no.~1, pp. 13\,795--13\,800, 2011.

\bibitem{ahmed2022technology}
H.~U. Ahmed, Y.~Huang, P.~Lu, and R.~Bridgelall, ``Technology developments and impacts of connected and autonomous vehicles: An overview,'' \emph{Smart Cities}, vol.~5, no.~1, pp. 382--404, 2022.

\bibitem{eskandarian2024advanced}
A.~Eskandarian, ``Advanced methods and algorithms for selected connected autonomous vehicles (cavs) benefits,'' \emph{IEEE Transactions on Intelligent Transportation Systems}, vol.~25, no.~4, pp. 405--442, 2024.

\bibitem{yuan2025bio}
F.~Yuan, Z.~Lin, Z.~Tian, B.~Chen, Q.~Zhou, C.~Yuan, H.~Sun, and Z.~Huang, ``Bio-inspired hybrid path planning for efficient and smooth robotic navigation: F. yuan et al.'' \emph{International Journal of Intelligent Robotics and Applications}, pp. 1--31, 2025.

\bibitem{tan2021human}
Z.~Tan, N.~Dai, Y.~Su, R.~Zhang, Y.~Li, D.~Wu, and S.~Li, ``Human--machine interaction in intelligent and connected vehicles: A review of status quo, issues, and opportunities,'' \emph{IEEE transactions on intelligent transportation systems}, vol.~23, no.~9, pp. 13\,954--13\,975, 2021.

\bibitem{ezzati2021interaction}
R.~Ezzati~Amini, C.~Katrakazas, A.~Riener, and C.~Antoniou, ``Interaction of automated driving systems with pedestrians: Challenges, current solutions, and recommendations for ehmis,'' \emph{Transport Reviews}, vol.~41, no.~6, pp. 788--813, 2021.

\bibitem{wang2024towards}
J.~Wang, L.~Su, S.~Han, D.~Song, and F.~Miao, ``Towards safe autonomy in hybrid traffic: Detecting unpredictable abnormal behaviors of human drivers via information sharing,'' \emph{ACM Transactions on Cyber-Physical Systems}, vol.~8, no.~2, pp. 1--25, 2024.

\bibitem{chen2024review}
Y.~Chen, ``Review of autonomous driving in unexpected events,'' in \emph{2024 International Conference on Intelligent Robotics and Automatic Control (IRAC)}.\hskip 1em plus 0.5em minus 0.4em\relax IEEE, 2024, pp. 45--51.

\bibitem{tian2025evaluating}
Z.~Tian, Z.~Lin, D.~Zhao, W.~Zhao, D.~Flynn, S.~Ansari, and C.~Wei, ``Evaluating scenario-based decision-making for interactive autonomous driving using rational criteria: A survey,'' \emph{arXiv preprint arXiv:2501.01886}, 2025.

\bibitem{yao2020path}
Q.~Yao, Z.~Zheng \emph{et~al.}, ``Path planning method with improved artificial potential field—a reinforcement learning perspective,'' \emph{IEEE access}, vol.~8, pp. 135\,513--135\,523, 2020.

\bibitem{triharminto2017local}
H.~H. Triharminto, O.~Wahyunggoro, T.~B. Adji, A.~Cahyadi, I.~Ardiyanto, and Iswanto, ``Local information using stereo camera in artificial potential field based path planning,'' \emph{IAENG International Journal of Computer Science}, vol.~44, no.~3, pp. 316--326, 2017.

\bibitem{triharminto2016novel}
H.~H. Triharminto, O.~Wahyunggoro \emph{et~al.}, ``A novel of repulsive function on artificial potential field for robot path planning,'' \emph{International Journal of Electrical and Computer Engineering}, vol.~6, no.~6, p. 3262, 2016.

\bibitem{li2025adaptive}
Q.~Li, Z.~Tian, X.~Wang, J.~Yang, and Z.~Lin, ``Adaptive field effect planner for safe interactive autonomous driving on curved roads,'' \emph{arXiv preprint arXiv:2504.14747}, 2025.

\bibitem{lofberg2003minimax}
J.~Löfberg, ``Minimax approaches to robust model predictive control,'' vol. 812, pp. 29--37, 2003.

\bibitem{glover2021h}
K.~Glover, ``H-infinity control,'' in \emph{Encyclopedia of systems and control}.\hskip 1em plus 0.5em minus 0.4em\relax Springer, 2021, pp. 896--902.

\bibitem{wu2021sliding}
L.~Wu, J.~Liu, S.~Vazquez, and S.~K. Mazumder, ``Sliding mode control in power converters and drives: A review,'' \emph{IEEE/CAA Journal of Automatica Sinica}, vol.~9, no.~3, pp. 392--406, 2021.

\bibitem{li2025efficient}
Q.~Li, Z.~Tian, X.~Wang, J.~Yang, and Z.~Lin, ``Efficient and safe planner for automated driving on ramps considering unsatisfication,'' \emph{arXiv preprint arXiv:2504.15320}, 2025.

\bibitem{tian2025risk}
Z.~Tian, Z.~Lin, D.~Zhao, C.~Anagnostopoulos, Q.~Wang, W.~Zhao, X.~Wang, and C.~Wei, ``A risk-aware spatial-temporal trajectory planning framework for autonomous vehicles using qp-mpc and dynamic hazard fields,'' \emph{arXiv preprint arXiv:2509.00643}, 2025.

\bibitem{browne2012survey}
C.~B. Browne \emph{et~al.}, ``A survey of monte carlo tree search methods,'' \emph{IEEE Transactions on Computational Intelligence and AI in games}, vol.~4, no.~1, pp. 1--43, 2012.

\bibitem{lenz2016tactical}
D.~Lenz, T.~Kessler, and A.~Knoll, ``Tactical cooperative planning for autonomous highway driving using monte-carlo tree search,'' in \emph{2016 IEEE Intelligent Vehicles Symposium (IV)}.\hskip 1em plus 0.5em minus 0.4em\relax IEEE, 2016, pp. 447--453.

\bibitem{lin2025multi}
Z.~Lin \emph{et~al.}, ``Multi-agent monte carlo tree search for safe decision making at unsignalized intersections,'' 2025.

\bibitem{lin2025safetyr}
Z.~Lin, S.~Liu, Z.~Tian, D.~Zhao, and J.~Lan, ``Safety-critical multi-agent mcts for mixed traffic coordination at unsignalized roundabout,'' \emph{arXiv preprint arXiv:2509.01856}, 2025.

\bibitem{xu2024recent}
T.~Xu, ``Recent advances in rapidly-exploring random tree: A review,'' \emph{Heliyon}, 2024.

\bibitem{li2021adaptive}
B.~Li and B.~Chen, ``An adaptive rapidly-exploring random tree,'' \emph{IEEE/CAA Journal of Automatica Sinica}, vol.~9, no.~2, pp. 283--294, 2021.

\bibitem{bhattacharyya2021hybrid}
R.~Bhattacharyya, S.~Jung, L.~A. Kruse, R.~Senanayake, and M.~J. Kochenderfer, ``A hybrid rule-based and data-driven approach to driver modeling through particle filtering,'' \emph{IEEE Transactions on Intelligent Transportation Systems}, vol.~23, no.~8, pp. 13\,055--13\,068, 2021.

\bibitem{xiao2021rule}
W.~Xiao, N.~Mehdipour, A.~Collin, A.~Y. Bin-Nun, E.~Frazzoli, R.~D. Tebbens, and C.~Belta, ``Rule-based optimal control for autonomous driving,'' in \emph{Proceedings of the ACM/IEEE 12th International Conference on Cyber-Physical Systems}, 2021, pp. 143--154.

\bibitem{bouchard2022rule}
F.~Bouchard, S.~Sedwards, and K.~Czarnecki, ``A rule-based behaviour planner for autonomous driving,'' in \emph{Proceeding of the International Joint Conference on Rules and Reasoning}.\hskip 1em plus 0.5em minus 0.4em\relax Springer, 2022, pp. 263--279.

\bibitem{9718218}
X.~Tang, B.~Huang, T.~Liu, and X.~Lin, ``Highway decision-making and motion planning for autonomous driving via soft actor-critic,'' \emph{IEEE Transactions on Vehicular Technology}, vol.~71, no.~5, pp. 4706--4717, 2022.

\bibitem{li2018estimating}
X.~Li, W.~Wang, and M.~Roetting, ``Estimating driver’s lane-change intent considering driving style and contextual traffic,'' \emph{IEEE Trans. Intell. Transp. Syst.}, vol.~20, no.~9, pp. 3258--3271, 2018.

\bibitem{hu2025applications}
J.~Hu \emph{et~al.}, ``Applications and effect evaluation of generative adversarial networks in semi-supervised learning,'' \emph{arXiv preprint arXiv:2505.19522}, 2025.

\bibitem{zhen2025balanced}
T.~Zhen \emph{et~al.}, ``Balanced exploration and attention-inspired decision making for autonomous driving,'' \emph{IEEE Transactions on Vehicular Technology}, 2025.

\bibitem{10214640}
G.~Chen, Y.~Zhang, and X.~Li, ``Attention-based highway safety planner for autonomous driving via deep reinforcement learning,'' \emph{IEEE Transactions on Vehicular Technology}, vol.~73, no.~1, pp. 162--175, 2024.

\bibitem{cai2021vision}
P.~Cai, H.~Wang \emph{et~al.}, ``Vision-based autonomous car racing using deep imitative reinforcement learning,'' \emph{IEEE Robotics and Automation Letters}, vol.~6, no.~4, pp. 7262--7269, 2021.

\bibitem{bacsar1998dynamic}
T.~Ba{\c{s}}ar and G.~J. Olsder, \emph{Dynamic noncooperative game theory}.\hskip 1em plus 0.5em minus 0.4em\relax SIAM, 1998.

\bibitem{8678699}
K.~Huang, C.~Zhou, Y.~Qin, and W.~Tu, ``A game-theoretic approach to cross-layer security decision-making in industrial cyber-physical systems,'' \emph{IEEE Transactions on Industrial Electronics}, vol.~67, no.~3, pp. 2371--2379, 2020.

\bibitem{kita2002game}
H.~Kita, K.~Tanimoto, and K.~Fukuyama, ``A game theoretic analysis of merging-giveway interaction: a joint estimation model,'' in \emph{Transportation and Traffic Theory in the 21st Century}.\hskip 1em plus 0.5em minus 0.4em\relax Emerald Group Publishing Limited, 2002, pp. 503--518.

\bibitem{liu2025data}
Y.~Liu, Z.~Tian, J.~Yang, and Z.~Lin, ``Data-driven evolutionary game-based model predictive control for hybrid renewable energy dispatch in autonomous ships,'' in \emph{2025 4th International Conference on New Energy System and Power Engineering (NESP)}.\hskip 1em plus 0.5em minus 0.4em\relax IEEE, 2025, pp. 482--490.

\bibitem{yu2018human}
H.~Yu, H.~E. Tseng, and R.~Langari, ``A human-like game theory-based controller for automatic lane changing,'' \emph{Transportation Research Part C: Emerging Technologies}, vol.~88, pp. 140--158, 2018.

\bibitem{zheng2025mean}
L.~Zheng, X.~Wang \emph{et~al.}, ``A mean-field-game-integrated mpc-qp framework for collision-free multi-vehicle control,'' \emph{Drones}, vol.~9, no.~5, p. 375, 2025.

\bibitem{zheng2025enhanced}
L.~Zheng \emph{et~al.}, ``Enhanced mean field game for interactive decision-making with varied stylish multi-vehicles,'' \emph{arXiv preprint arXiv:2509.00981}, 2025.

\bibitem{claussmann2019review}
L.~Claussmann, M.~Revilloud, D.~Gruyer, and S.~Glaser, ``A review of motion planning for highway autonomous driving,'' \emph{IEEE Transactions on Intelligent Transportation Systems}, vol.~21, no.~5, pp. 1826--1848, 2019.

\bibitem{milanes2010automated}
V.~Milan{\'e}s, J.~Godoy, J.~Villagr{\'a}, and J.~P{\'e}rez, ``Automated on-ramp merging system for congested traffic situations,'' \emph{IEEE Transactions on Intelligent Transportation Systems}, vol.~12, no.~2, pp. 500--508, 2010.

\bibitem{zhang2025evaluation}
C.~Zhang, C.~Tian, T.~Han, H.~Li, Y.~Feng, Y.~Chen, R.~W. Proctor, and J.~Zhang, ``Evaluation of an infrastructure-based warning system: A case study on roundabout driving behaviors,'' \emph{IEEE Transactions on Intelligent Transportation Systems}, 2025.

\bibitem{xu2019cooperative}
H.~Xu, Y.~Zhang, L.~Li, and W.~Li, ``Cooperative driving at unsignalized intersections using tree search,'' \emph{IEEE Transactions on Intelligent Transportation Systems}, vol.~21, no.~11, pp. 4563--4571, 2019.

\bibitem{betz2022autonomous}
J.~Betz, H.~Zheng, A.~Liniger, U.~Rosolia, P.~Karle, M.~Behl, V.~Krovi, and R.~Mangharam, ``Autonomous vehicles on the edge: A survey on autonomous vehicle racing,'' \emph{IEEE Open Journal of Intelligent Transportation Systems}, vol.~3, pp. 458--488, 2022.

\bibitem{chen2023safe}
J.~Chen, C.~Zhao, S.~Jiang, X.~Zhang, Z.~Li, and Y.~Du, ``Safe, efficient, and comfortable autonomous driving based on cooperative vehicle infrastructure system,'' \emph{International journal of environmental research and public health}, vol.~20, no.~1, p. 893, 2023.

\bibitem{cominetti2010payoff}
R.~Cominetti, E.~Melo, and S.~Sorin, ``A payoff-based learning procedure and its application to traffic games,'' \emph{Games and Economic Behavior}, vol.~70, no.~1, pp. 71--83, 2010.

\bibitem{kreps1989nash}
D.~M. Kreps, ``Nash equilibrium,'' in \emph{Game theory}.\hskip 1em plus 0.5em minus 0.4em\relax Springer, 1989, pp. 167--177.

\bibitem{mariani2021coordination}
S.~Mariani, G.~Cabri, and F.~Zambonelli, ``Coordination of autonomous vehicles: taxonomy and survey,'' \emph{ACM Computing Surveys (CSUR)}, vol.~54, no.~1, pp. 1--33, 2021.

\bibitem{hang2021cooperative}
P.~Hang, C.~Lv, C.~Huang, Y.~Xing, and Z.~Hu, ``Cooperative decision making of connected automated vehicles at multi-lane merging zone: A coalitional game approach,'' \emph{IEEE Transactions on Intelligent Transportation Systems}, vol.~23, no.~4, pp. 3829--3841, 2021.

\bibitem{ding2019multivehicle}
N.~Ding, X.~Meng, W.~Xia, D.~Wu, L.~Xu, and B.~Chen, ``Multivehicle coordinated lane change strategy in the roundabout under internet of vehicles based on game theory and cognitive computing,'' \emph{IEEE Transactions on Industrial Informatics}, vol.~16, no.~8, pp. 5435--5443, 2019.

\bibitem{jain2022bayesian}
N.~Jain and S.~Mittal, ``Bayesian nash equilibrium based gaming model for eco-safe driving,'' \emph{Journal of King Saud University-Computer and Information Sciences}, vol.~34, no.~9, pp. 7482--7493, 2022.

\bibitem{faillo2013roles}
M.~Faillo, A.~Smerilli, and R.~Sugden, ``The roles of level-k and team reasoning in solving coordination games,'' \emph{Cognitive and Experimental Economics Laboratory Working Paper}, no. 6-13, 2013.

\bibitem{huang2025fair}
Z.~Huang, C.~Hao, W.~Zhan, J.~Ma, and M.~Tomizuka, ``Fair play in the fast lane: Integrating sportsmanship into autonomous racing systems,'' \emph{arXiv preprint arXiv:2503.03774}, 2025.

\bibitem{Kim2014}
C.~Kim and R.~Langari, ``Game theory based autonomous vehicles operation,'' \emph{International Journal of Vehicle Design}, vol.~65, no.~4, pp. 360--383, 2014.

\bibitem{Talebpour2015}
A.~Talebpour, H.~S. Mahmassani, and S.~H. Hamdar, ``Modeling lane-changing behavior in a connected environment: A game theory approach,'' in \emph{Transport. Res. Procedia (Proc. IEEE ITSC Workshop)}, vol.~7, 2015, pp. 420--440.

\bibitem{Wang2015}
M.~Wang, S.~P. Hoogendoorn, W.~Daamen, B.~van Arem, and R.~Happee, ``Game theoretic approach for predictive lane-changing and car-following control,'' \emph{Transportation Research Part C: Emerging Technologies}, vol.~58, pp. 73--92, 2015.

\bibitem{Sadigh2018AR}
D.~Sadigh, N.~Landolfi, S.~S. Sastry, S.~A. Seshia, and A.~D. Dragan, ``Planning for cars that coordinate with people: leveraging effects on human actions for planning and active information gathering,'' \emph{Autonomous Robots}, vol.~42, no.~7, pp. 1405--1426, 2018.

\bibitem{Yoo2012Highway}
J.~H. Yoo and R.~Langari, ``Stackelberg game based model of highway driving,'' in \emph{Proc. ASME Dynamic Systems and Control Conf. (DSCC)}, 2012, pp. 499--508.

\bibitem{Hang2020Noncoop}
P.~Hang, C.~Lv, Z.~Huang, Z.~Hu, and Y.~Xing, ``Human-like decision making for autonomous driving: A noncooperative game theoretic approach,'' \emph{IEEE Trans. Intelligent Transportation Systems}, vol.~22, no.~4, pp. 2076--2087, 2021.

\bibitem{Coskun2019}
S.~Coskun, Q.~Zhang, and R.~Langari, ``Receding horizon markov game autonomous driving strategy,'' in \emph{Proc. American Control Conf. (ACC)}, 2019, pp. 1367--1374.

\bibitem{hang2020integrated}
P.~Hang, C.~Lv, C.~Huang, J.~Cai, Z.~Hu, and Y.~Xing, ``An integrated framework of decision making and motion planning for autonomous vehicles considering social behaviors,'' \emph{IEEE transactions on vehicular technology}, vol.~69, no.~12, pp. 14\,458--14\,469, 2020.

\bibitem{yan2018highway}
Z.~Yan, J.~Wang, and Y.~Zhang, ``A game-theoretical approach to driving decision making in highway scenarios,'' in \emph{2018 IEEE Intelligent Vehicles Symposium (IV)}, 2018, pp. 1221--1226.

\bibitem{higashi2003particle}
N.~Higashi and H.~Iba, ``Particle swarm optimization with gaussian mutation,'' in \emph{Proceedings of the 2003 IEEE Swarm Intelligence Symposium. SIS'03 (Cat. No. 03EX706)}.\hskip 1em plus 0.5em minus 0.4em\relax IEEE, 2003, pp. 72--79.

\bibitem{shang2023emergency}
X.~Shang and A.~Eskandarian, ``Emergency collision avoidance and mitigation using model predictive control and artificial potential function,'' \emph{IEEE Transactions on Intelligent Vehicles}, vol.~8, no.~5, pp. 3458--3472, 2023.

\bibitem{liu2025aco}
Y.~Liu \emph{et~al.}, ``An aco-mpc framework for energy-efficient and collision-free path planning in autonomous maritime navigation,'' in \emph{2025 8th International Conference on Advanced Algorithms and Control Engineering (ICAACE)}.\hskip 1em plus 0.5em minus 0.4em\relax IEEE, 2025, pp. 344--354.

\bibitem{Schwarting2019}
W.~Schwarting, A.~Pierson, J.~Alonso-Mora, S.~Karaman, and D.~Rus, ``Social behavior for autonomous vehicles,'' \emph{Proceedings of the National Academy of Sciences (PNAS)}, vol. 116, no.~50, pp. 24\,972--24\,978, 2019.

\bibitem{Liu2024PotentialGame}
M.~Liu, H.~E. Tseng, D.~Filev, A.~Girard, and I.~Kolmanovsky, ``Safe and human-like autonomous driving: A predictor-corrector potential game approach,'' \emph{IEEE Trans. Control Systems Technology}, vol.~32, no.~2, pp. 834--848, 2024.

\bibitem{murphy2011measuring}
R.~O. Murphy, K.~A. Ackermann, and M.~J. Handgraaf, ``Measuring social value orientation,'' \emph{Judgment and Decision making}, vol.~6, no.~8, pp. 771--781, 2011.

\bibitem{gong2022modeling}
B.~Gong, F.~Wang, C.~Lin, and D.~Wu, ``Modeling hdv and cav mixed traffic flow on a foggy two-lane highway with cellular automata and game theory model,'' \emph{Sustainability}, vol.~14, no.~10, p. 5899, 2022.

\bibitem{Lopez2022}
V.~G. Lopez, F.~L. Lewis, M.~Liu, Y.~Wan, S.~Nageshrao, and D.~Liu, ``Game-theoretic lane-changing decision making and payoff learning for autonomous vehicles,'' \emph{IEEE Trans. Vehicular Technology}, vol.~71, no.~4, pp. 3609--3620, 2022.

\bibitem{vogel2003comparison}
K.~Vogel, ``A comparison of headway and time to collision as safety indicators,'' \emph{Accident analysis \& prevention}, vol.~35, no.~3, pp. 427--433, 2003.

\bibitem{sidaway1996time}
B.~Sidaway, M.~Fairweather, H.~Sekiya, and J.~Mcnitt-Gray, ``Time-to-collision estimation in a simulated driving task,'' \emph{Human factors}, vol.~38, no.~1, pp. 101--113, 1996.

\bibitem{karimi2023deep}
S.~Karimi, ``Deep reinforcement learning and game theoretic monte carlo decision process for safe and efficient lane change maneuver and speed management,'' 2023.

\bibitem{huang2024highway}
P.~Huang, H.~Ding, Z.~Sun, and H.~Chen, ``A game-based hierarchical model for mandatory lane change of autonomous vehicles,'' \emph{IEEE Transactions on Intelligent Transportation Systems}, vol.~25, no.~9, pp. 11\,256--11\,268, 2024.

\bibitem{Jain2024Survey}
G.~Jain, A.~Kumar, and S.~A. Bhat, ``Recent developments of game theory and reinforcement learning approaches: A systematic review,'' \emph{IEEE Access}, vol.~12, p. 99990011, 2024.

\bibitem{Di2021Survey}
X.~Di and R.~Shi, ``A survey on autonomous vehicle control in the era of mixed-autonomy: From physics-based to ai-guided driving policy learning,'' \emph{Transportation Research Part C: Emerging Technologies}, vol. 125, p. 103008, 2021.

\bibitem{Toghi2022Altruism}
B.~Toghi, R.~Valiente, D.~Sadigh, R.~Pedarsani, and Y.~P. Fallah, ``Social coordination and altruism in autonomous driving,'' \emph{IEEE Trans. Intelligent Transportation Systems}, vol.~23, no.~10, pp. 24\,791--24\,804, 2022.

\bibitem{Nishi2019MergingRL}
T.~Nishi, P.~Doshi, and D.~Prokhorov, ``Merging in congested freeway traffic using multi-policy decision making and passive actor-critic learning,'' \emph{IEEE Trans. Intelligent Vehicles}, vol.~4, no.~2, pp. 287--297, 2019.

\bibitem{zhou2024highway}
X.~Zhou, Z.~Peng, Y.~Xie, M.~Liu, and J.~Ma, ``Game-theoretic driver modeling and decision-making for autonomous driving with temporal-spatial attention-based deep q-learning,'' \emph{IEEE Transactions on Intelligent Vehicles}, pp. 1--17, 2024.

\bibitem{Wei2022Merging}
C.~Wei, Y.~He, H.~Tian, and Y.~Lv, ``Game theoretic merging behavior control for autonomous vehicles at highway on-ramps,'' \emph{IEEE Trans. Intelligent Transportation Systems}, vol.~23, no.~11, pp. 21\,127--21\,136, 2022.

\bibitem{Yoo2013Merging}
J.~H. Yoo and R.~Langari, ``A stackelberg game theoretic driver model for merging,'' in \emph{Proc. ASME Dynamic Systems and Control Conf.}, 2013, p. V002T30A003.

\bibitem{Ji2021Stackelberg}
K.~Ji, M.~Orsag, and K.~Han, ``Lane-merging strategy for a self-driving car in dense traffic using the stackelberg game approach,'' \emph{Electronics}, vol.~10, no.~7, p. 894, 2021.

\bibitem{Wang2021Ramp}
H.~Wang, W.~Wang, S.~Yuan, X.~Li, and L.~Sun, ``On social interactions of merging behaviors at highway on-ramps in congested traffic,'' \emph{IEEE Trans. Intelligent Transportation Systems}, vol.~23, no.~11, pp. 11\,237--11\,248, 2022.

\bibitem{li2024merging}
W.~Li, F.~Qiu, L.~Li, Y.~Zhang, and K.~Wang, ``Simulation of vehicle interaction behavior in merging scenarios: A deep maximum entropy-inverse reinforcement learning method combined with game theory,'' \emph{IEEE Transactions on Intelligent Vehicles}, vol.~9, no.~1, pp. 1079--1093, 2024.

\bibitem{sze2017efficient}
V.~Sze, Y.-H. Chen, T.-J. Yang, and J.~S. Emer, ``Efficient processing of deep neural networks: A tutorial and survey,'' \emph{Proceedings of the IEEE}, vol. 105, no.~12, pp. 2295--2329, 2017.

\bibitem{Liu2022ForcedMerge}
K.~Liu, N.~Li, H.~E. Tseng, I.~Kolmanovsky, and A.~Girard, ``Interaction-aware trajectory prediction and planning for autonomous vehicles in forced merge scenarios,'' \emph{IEEE Trans. Intelligent Transportation Systems}, vol.~24, no.~1, pp. 474--488, 2023.

\bibitem{chen2023game}
W.~Chen, G.~Ren, Q.~Cao, J.~Song, Y.~Liu, and C.~Dong, ``A game-theory-based approach to modeling lane-changing interactions on highway on-ramps: Considering the bounded rationality of drivers,'' \emph{Mathematics}, vol.~11, no.~2, p. 402, 2023.

\bibitem{Rahmati2017LeftTurn}
Y.~Rahmati and A.~Talebpour, ``Towards a collaborative connected, automated driving environment: A game theory based decision framework for unprotected left turn maneuvers,'' in \emph{Proc. IEEE Intelligent Vehicles Symposium (IV)}, 2017, pp. 1316--1321.

\bibitem{Rahmati2020HumanVeh}
Y.~Rahmati, A.~Talebpour, A.~Mittal, and J.~Fishelson, ``Game theory-based framework for modeling human-vehicle interactions on the road,'' \emph{Transportation Research Record}, vol. 2674, no.~5, pp. 701--713, 2020.

\bibitem{Li2024IRLMerge}
W.~Li, F.~Qiu, L.~Li, Y.~Zhang, and K.~Wang, ``Simulation of vehicle interaction behavior in merging scenarios: A deep maximum entropy inverse reinforcement learning method combined with game theory,'' \emph{IEEE Trans. Intelligent Vehicles}, vol.~9, no.~2, pp. 1079--1093, 2024.

\bibitem{Li2018TCST}
N.~Li, D.~W. Oyler, M.~Zhang, Y.~Yildiz, I.~Kolmanovsky, and A.~R. Girard, ``{Game theoretic modeling of driver and vehicle interactions for verification and validation of autonomous vehicle control systems},'' \emph{IEEE Transactions on Control Systems Technology}, vol.~26, no.~5, pp. 1782--1797, 2018.

\bibitem{Tian2018Roundabout}
R.~Tian, S.~Li, N.~Li, I.~Kolmanovsky, and t.~A.~Girard, ``Adaptive game-theoretic decision making for autonomous vehicle control at roundabouts,'' in \emph{Proc. IEEE Conf. Decision and Control (CDC)}, 2018, pp. 321--326.

\bibitem{Li2019CDC}
S.~Li, N.~Li, A.~Girard, and I.~Kolmanovsky, ``{Decision making in dynamic and interactive environments based on cognitive hierarchy theory, Bayesian inference, and predictive control},'' in \emph{Proc.\ 58th IEEE Conf.\ on Decision and Control (CDC)}, Nice, France, 2019, pp. 2181--2187.

\bibitem{brunner2013connection}
N.~Brunner and N.~Linden, ``Connection between bell nonlocality and bayesian game theory,'' \emph{Nature communications}, vol.~4, no.~1, p. 2057, 2013.

\bibitem{Ding2020}
C.~Ding, H.~Wang, Y.~Qin, and K.~Li, ``{Coordinated multi-vehicle lane-changing strategy in roundabouts using a Stackelberg game approach},'' in \emph{Proc.\ IEEE Intelligent Vehicles Symposium (IV)}, Las Vegas, NV, USA, 2020.

\bibitem{Li2022IMechE}
D.~Li, G.~Liu, and B.~Xiao, ``{Human-like driving decision at unsignalized intersections based on game theory},'' \emph{Proc.\ IMechE Part D: Journal of Automobile Engineering}, vol. 236, no.~12, pp. 2890--2902, 2022.

\bibitem{Hang2022}
P.~Hang, Y.~Zhang, N.~de~Boer, and C.~Lv, ``{Conflict resolution for connected automated vehicles at unsignalized roundabouts considering personalized driving behaviours},'' \emph{Green Energy and Intelligent Transportation}, vol.~1, no.~4, p. 100003, 2022.

\bibitem{Banjanovic2016}
L.~Banjanovic-Mehmedovic, E.~Halilovic, I.~Bosankic, M.~Kantardzic, and S.~Kasapovic, ``Autonomous vehicle-to-vehicle (v2v) decision making in roundabout using game theory,'' \emph{Int. J. Advanced Computer Science and Applications}, vol.~7, no.~5, pp. 244--250, 2016.

\bibitem{Fisac2019}
J.~F. Fisac, E.~Bronstein, E.~Stefansson, D.~Sadigh, S.~S. Sastry, and C.~J. Tomlin, ``Hierarchical game-theoretic planning for autonomous vehicles,'' in \emph{Proc. IEEE Int. Conf. Robotics and Automation (ICRA)}, 2019, pp. 9590--9596.

\bibitem{Shu2023HumanInspired}
K.~Shu, R.~V. Mehrizi, S.~Li, M.~Pirani, and A.~Khajepour, ``Human inspired autonomous intersection handling using game theory,'' \emph{IEEE Trans. Intelligent Transportation Systems}, vol.~24, no.~11, pp. 11\,360--11\,371, 2023.

\bibitem{Wang2021KBTree}
H.~Wang, J.~Xi, and Y.~Ma, ``Kb-tree: Learnable and continuous monte-carlo tree search for motion planning in autonomous driving,'' in \emph{Proc. IEEE/RSJ Int. Conf. Intelligent Robots and Systems (IROS)}, 2021, pp. 4543--4549.

\bibitem{Li2020Unsignalized}
N.~Li, Y.~Yao, I.~Kolmanovsky, E.~Atkins, and A.~R. Girard, ``Game-theoretic modeling of multi-vehicle interactions at uncontrolled intersections,'' \emph{IEEE Trans. Intelligent Transportation Systems}, vol.~23, no.~3, pp. 1428--1442, 2022.

\bibitem{Tian2020Unsignalized}
R.~Tian, N.~Li, I.~Kolmanovsky, Y.~Yildiz, and t.~A.~Girard, ``Game-theoretic modeling of traffic in unsignalized intersection networks for autonomous vehicle control verification and validation,'' \emph{IEEE Trans. Intelligent Transportation Systems}, vol.~23, no.~3, pp. 2211--2226, 2022.

\bibitem{Nan2022Nash}
J.~Nan, W.~Deng, and B.~Zheng, ``Intention prediction and mixed strategy nash equilibrium-based decision-making framework for autonomous driving in uncontrolled intersections,'' \emph{IEEE Trans. Vehicular Technology}, vol.~71, no.~10, pp. 10\,316--10\,326, 2022.

\bibitem{Jia2023Switch}
S.~Jia, Y.~Zhang, X.~Li, X.~Na, and t.~Y.~Wang, ``Interactive decision-making with switchable game modes for automated vehicles at intersections,'' \emph{IEEE Trans. Intelligent Transportation Systems}, vol.~24, no.~11, pp. 11\,785--11\,799, 2023.

\bibitem{Zhao2021Yield}
X.~Zhao, Y.~Tian, and J.~Sun, ``Yield or rush? social-preference-aware driving interaction modeling using a game-theoretic framework,'' in \emph{Proc. IEEE Int. Intelligent Transportation Systems Conf. (ITSC)}, 2021, pp. 453--459.

\bibitem{Qin2024Review}
Z.~Qin, A.~Ji, Z.~Sun, G.~Wu, and t.~P.~Hao, ``Game theoretic application to intersection management: a literature review,'' \emph{IEEE Trans. Intelligent Vehicles}, vol.~9, no.~1, pp. 68--86, 2024.

\bibitem{yuan2022}
M.~Yuan, J.~Shan, and K.~Mi, ``Deep reinforcement learning based game-theoretic decision-making for autonomous vehicles,'' \emph{IEEE Robotics and Automation Letters}, vol.~7, no.~2, pp. 818--825, 2022.

\bibitem{Negash2023Review}
N.~Negash and J.~Yang, ``Driver behavior modeling toward autonomous vehicles: comprehensive review,'' \emph{IEEE Access}, vol.~11, pp. 22\,788--22\,821, 2023.

\bibitem{zhang2017energy}
Q.~Zhang, M.~Lin, L.~T. Yang, Z.~Chen, and P.~Li, ``Energy-efficient scheduling for real-time systems based on deep q-learning model,'' \emph{IEEE transactions on sustainable computing}, vol.~4, no.~1, pp. 132--141, 2017.

\bibitem{Wang2019RSS}
M.~Wang, Z.~Wang, J.~Talbot, J.~C. Gerdes, and M.~Schwager, ``Game theoretic planning for self-driving cars in competitive scenarios,'' in \emph{Proc. Robotics: Science and Systems (RSS)}, 2019, pp. 1--9.

\bibitem{Notomista2020CBF}
G.~Notomista, M.~Wang, M.~Schwager, and M.~Egerstedt, ``Enhancing game-theoretic autonomous car racing using control barrier functions,'' in \emph{IEEE International Conference on Robotics and Automation (ICRA)}, 2020, pp. 5390--5396.

\bibitem{Liniger2019Game}
A.~Liniger and J.~Lygeros, ``A non-cooperative game approach to autonomous racing,'' \emph{IEEE Transactions on Control Systems Technology}, vol.~28, no.~3, pp. 884--897, 2019.

\bibitem{Jung2023gameMPC}
C.~Jung, S.~Lee, H.~Seong, A.~Finazzi, and H.~Shim, ``Game-theoretic model predictive control with data-driven identification of vehicle model for head-to-head autonomous racing,'' in \emph{ICRA Workshop on AI-enabled Autonomous Racing}, 2023, arXiv:2210.10250.

\bibitem{Thakkar2024FR}
R.~S. Thakkar, A.~S. Samyal, D.~Fridovich-Keil, Z.~Xu, and U.~Topcu, ``Hierarchical control for head-to-head autonomous racing,'' \emph{Field Robotics}, vol.~4, pp. 46--69, 2024.

\bibitem{Huang2025sports}
Z.~Huang, C.~Hao, W.~Zhan, J.~Ma, and M.~Tomizuka, ``Fair play in the fast lane: Integrating sportsmanship into autonomous racing systems,'' \emph{arXiv preprint arXiv:2503.03774}, 2025.

\bibitem{Jia2023RAPID}
Y.~Jia, M.~Bhatt, and N.~Mehr, ``{RAPID}: Autonomous multi-agent racing using constrained potential dynamic games,'' in \emph{Proc. IEEE Conf. on Decision and Control (CDC)}, 2023, arXiv:2305.00579.

\bibitem{boggs1995sequential}
P.~T. Boggs and J.~W. Tolle, ``Sequential quadratic programming,'' \emph{Acta numerica}, vol.~4, pp. 1--51, 1995.

\bibitem{Rowold2024IV}
M.~Rowold, A.~Langmann, B.~Lohmann, and J.~Betz, ``Open-loop and feedback {Nash} trajectories for competitive racing with ilqgames,'' \emph{Proc. IEEE Intelligent Vehicles Symposium (IV)}, 2024, arXiv:2402.01918.

\bibitem{Zhu2024GNE}
E.~L. Zhu and F.~Borrelli, ``A sequential quadratic programming approach to the solution of open-loop generalized {Nash} equilibria for autonomous racing,'' \emph{Proc. IEEE Intelligent Vehicles Symposium (IV)}, 2024, arXiv:2404.00186.

\end{thebibliography}

\vfill

\end{document}